%% file: main.tex
\documentclass{article}

     \usepackage[preprint]{neurips_2025}

\PassOptionsToPackage{table, svgnames, x11names}{xcolor}
\usepackage[utf8]{inputenc} 
\usepackage[T1]{fontenc}    
\usepackage{hyperref}       
\usepackage{url}            
\usepackage{booktabs}       
\usepackage{amsfonts}       
\usepackage{nicefrac}       
\usepackage{microtype}      
\usepackage{algorithm}
\usepackage{algpseudocode}
\usepackage{amssymb}
\usepackage{amsthm}
\usepackage{amsmath}
\usepackage{paralist}
\usepackage{setspace}
\usepackage{graphicx}
\usepackage{natbib}
\usepackage{subcaption}
\usepackage{multirow}
\usepackage{multicol}
\usepackage{makecell}
\usepackage{xspace}
\usepackage{svg}
\usepackage[table]{xcolor}

\newtheorem{lem}{Lemma}[section]
\newtheorem{defi}{Definition}[section]
\newtheorem{theorem}{Theorem}[section]
\newcommand{\round}[1]{\ensuremath{\lfloor#1\rceil}}

\newcommand{\sysname}{\texttt{FireQ}\xspace}

\title{\sysname: Fast INT4-FP8 Kernel and RoPE-aware Quantization for LLM Inference Acceleration}

\author{%
  Daehyeon Baek$^{*}$, Jieun Choi$^{*}$, Jimyoung Son\thanks{The first three authors contribute equally and are listed alphabetically. $^\dag$Corresponding author.} \\
  {\textbf {Kyungmin Bin, Seungbeom Choi, Kihyo Moon, Minsung Jang, Hyojung Lee$^\dag$}}\\
  Cloud Research Team, Samsung SDS \\
}

\begin{document}

\maketitle

\input{sections/00-abstract}
\input{sections/01-intro}
\input{sections/02-background}
\input{sections/03-method}
\input{sections/04-experiments}
\input{sections/05-related_work}

\input{sections/06-conclusions}

\bibliographystyle{ACM-Reference-Format}
\bibliography{reference}

\input{sections/07-appendix}

\end{document}

%% file: sections/00-abstract.tex
\begin{abstract}
As large language models become increasingly prevalent, memory bandwidth constraints significantly limit inference throughput, motivating post-training quantization (PTQ). In this paper, we propose \sysname, a co-designed PTQ framework and an INT4-FP8 matrix multiplication kernel that accelerates LLM inference across all linear layers. Specifically, \sysname quantizes linear layer weights and key-values to INT4, and activations and queries to FP8, significantly enhancing throughput. Additionally, we introduce a three-stage pipelining for the prefill phase, which modifies the FlashAttention-3 kernel, effectively reducing time-to-first-token in the prefill phase. To minimize accuracy loss from quantization, we develop novel outlier smoothing techniques tailored separately for linear and attention layers. In linear layers, we explicitly use per-tensor scaling to prevent underflow caused by the FP8 quantization scaling factor of INT4 quantization, and channel-wise scaling to compensate for INT4’s coarse granularity. In attention layers, we address quantization challenges posed by rotary positional embeddings (RoPE) by combining pre-RoPE and post-RoPE scaling strategies. \sysname significantly outperforms state-of-the-art methods, achieving 1.68x faster inference in feed-forward network layers on Llama2-7B and 1.26x faster prefill phase performance on Llama3-8B compared to QServe, with negligible accuracy loss. Code is available at: \href{https://github.com/llm-fireq/fireq}{\color{blue}{FireQ.git}}.

\end{abstract}


%% file: sections/01-intro.tex
\section{Introduction}

As model sizes and input sequence lengths increase, inference throughput is increasingly bottlenecked by memory bandwidth.~\citep{skhynix-aim, memory-wall} Post-training quantization (PTQ) addresses this bottleneck by reducing memory footprint, thereby supporting larger batch sizes and longer sequences. However, quantization can introduce accuracy loss due to numerical instabilities, particularly from outlier values causing underflow or overflow in low-bit quantization. While existing outlier mitigation methods, such as rotation and scaling, help preserve accuracy, they often incur significant inference-time overhead. 

To overcome these limitations, we propose \sysname, a fast INT4-FP8 general matrix multiplication (GEMM) kernel explicitly optimized for accelerating quantized model inference. \sysname integrates effectively with quantization strategies for linear and attention layers: linear layer quantization addresses underflow issues caused by FP8 scaling factors in INT4 quantization, while attention layer quantization employs the rotary positional embeddings (RoPE)~\citep{su2021roformer}-aware scaling to enhance accuracy.

To accelerate linear layer computations, \sysname employs a specialized INT4-FP8 GEMM kernel optimized for our mixed-precision quantization strategy. Specifically, weights and activations are quantized to INT4 and FP8 (W4A8-FP), respectively, with FP8 scaling factors. Our kernel leverages CUDA cores to perform in-register INT4-to-FP8 dequantization, significantly improving throughput. To address underflow risks caused by the FP8 scaling factor, we implement per-tensor scaling. Additionally, channel-wise absmean scaling reduces inter-channel variance, compensating for INT4 quantization's coarse granularity, maintaining accuracy without added computational cost.

For attention layers, \sysname quantizes keys and values to INT4 and queries to FP8 (KV4Q8-FP). INT4 quantization of the key matrix presents unique challenges due to pronounced outliers compared to queries and values~\citep{lin2024qserve}. Additionally, RoPE introduces nonlinear transformations, complicating standard quantization techniques. To address this, we adopt post-RoPE quantization, reducing computational overhead relative to pre-RoPE methods, which require additional quantization and dequantization steps around RoPE. We further introduce a two-stage RoPE-aware outlier smoothing strategy: pre-RoPE normalization handles stable channel pairs, and targeted post-RoPE scaling addresses outlier channels. This method effectively preserves quantization accuracy without compromising throughput.

To accelerate the prefill phase of attention mechanisms, \sysname modifies FlashAttention-3~\citep{shah2024flashattention} by employing a producer-consumer warpgroup structure with three-stage pipelining. The producer warpgroup asynchronously loads query, key, and value matrices into shared memory, transposing the value matrix for optimal computation. The consumer warpgroup operates in three overlapping stages: (i) query-key multiplication with FP16 accumulation, (ii) softmax computation in FP16 and FP8 quantization, and (iii) value aggregation using FP8-quantized softmax outputs, with FP32 accumulation reduced to BF16. This pipeline effectively enhances hardware utilization and significantly reduces time-to-first-token (TTFT) for the prefill phase in inference.

We evaluate \sysname on H100 GPUs and compare its performance with state-of-the-art frameworks, including QServe~\citep{lin2024qserve}, Atom~\citep{zhao2024atom}, QuaRot~\citep{ashkboos2024quarot}, and TensorRT-LLM (W4A8-FP, AWQ)~\citep{tensorrt-llm, lin2024awq}. Our results demonstrate significant throughput improvements: \sysname achieves 1.8× higher throughput than QServe and 1.24× higher throughput than TensorRT-LLM for feed-forward network layers on the Llama2-7B model (batch size 16). For the Llama3-8B model with batch size 16 and sequence length 1024, \sysname also delivers a 1.26× speedup over QServe in the prefill phase. Despite these substantial throughput gains, \sysname maintains zero-shot accuracy comparable to existing frameworks, demonstrating an effective balance between inference efficiency and accuracy. Our ablation study further validates that the proposed quantization outlier smoothing strategies significantly contribute to reducing accuracy loss.


%% file: sections/02-background.tex
\section{Background}\label{sec:background}

\subsection{Post-training Quantization}

Post-training quantization (PTQ) encodes model weights and activations into lower-precision formats to reduce memory usage and computational overhead, without requiring model retraining. Specifically, symmetric uniform $b$-bit integer quantization converts a real-valued tensor $\mathbf{x}\in\mathbb{R}^n$ to integer values $\hat{\mathbf{x}}\in\{-2^{b-1},\dots,2^{b-1}-1\}^n$ using a scaling factor $\sigma$:
\begin{equation}
\sigma = \frac{\max_i |x_i|}{2^{b-1}-1},\quad \hat{x}_i = \round{\frac{x_i}{\sigma}}, \quad x_i^{(q)} \approx \sigma\cdot\hat{x}_i,    
\end{equation}
where $x_i^{(q)}$ denotes the dequantized approximation. In \sysname, weights in linear layers and key-value (KV) matrices in attention layers are quantized to INT4 with FP8 scaling factors. Activations, on the other hand, are quantized directly into FP8 format (e.g., E4M3) using a BF16 scaling factor $\beta$:
\begin{equation}
\hat{y}_i = \text{FP8}\left({y_i}/{\beta}\right), \quad y_i^{(q)} \approx \beta\cdot\hat{y}_i.    
\end{equation}
Quantization granularity differs by component: weights in linear layers use per-channel group-wise symmetric quantization, while keys and values in attention layers apply per-token quantization, and queries utilize FP8 precision.

\subsection{Challenges in LLM Quantization}
Quantizing LLMs introduces unique challenges due to their scaling, sensitivity to quantization noise, and complex transformer-based architecture. LLM inference operates autoregressively, generating tokens through iterative transformer blocks consisting of linear layers and multi-head attention layers:

\smallskip
\noindent \textbf{Linear layers.}
Linear layers perform general matrix multiplication between input activations and pre-trained weights, which covers up, down, gate projections in the feed-forward network (FFN) block, and query-key-value (QKV) generation, output projection in the multi-head attention (MHA) block, as shown in Figure~\ref{fig:overall_quantization}. A typical linear operation is defined as:
\begin{equation}
   \mathbf{Y} =  \mathbf{X}\mathbf{W}^{\text{T}},
\end{equation}
where $\mathbf{X}$ and $\mathbf{Y}$ are the input and output matrices, respectively, and $\mathbf{W} \in \mathbb{R}^{N_\text{out} \times N_\text{in}}$ represents weights for input $N_\text{in}$ and output $N_\text{out}$ dimension.

\smallskip
\noindent \textbf{Attention layer.} Attention layer calculates attention scores between query (Q) and key (K) vectors and computes a weighted sum of value (V) vectors. Given query, key, and value matrices $\mathbf{Q}, \mathbf{K} ,\mathbf{V} \in \mathbb{R}^{N \times d}$ for a single head, the attention output $\mathbf{O}$ is computed as:
    \begin{equation}
        \mathbf{S} = \frac{1}{\sqrt{d}}{\mathbf{Q}\mathbf{K}^{\text{T}}} \in \mathbb{R}^{N \times N}, \quad \mathbf{P} = \text{softmax}(\mathbf{S}) \in \mathbb{R}^{N \times N}, \quad \mathbf{O} = \mathbf{PV} \in \mathbb{R}^{N \times d}.
    \end{equation} 
Outputs from multiple heads are concatenated and projected through an output layer.

\noindent \textbf{Rotary positional embeddings (RoPE).} 
RoPE injects position information into query and key vectors after their linear projection, enabling position-aware attention. Given a query or key matrix $\mathbf{x}^t \in \mathbb{R}^{N \times d}$ at position $t$, RoPE applies a 2D rotation $R(\theta_i^t)$ to pairs of channel $(\mathbf{x}^t_i, \mathbf{x}^t_j)$, where $j = i+ d/2$ and $i = 0, \dots, d/2 -1$. The transformed vector $(\mathbf{x}^t_i, \mathbf{x}^t_{j})$ are computed by:
\begin{equation}
        \left(\tilde{\mathbf{x}}_i^t, \tilde{\mathbf{x}}_{j}^t \right) = 
        \left({\mathbf{x}}_i^t, {\mathbf{x}}_{j}^t\right) \cdot R(\theta_i^{t}), \quad \text{and} \quad
        R(\theta^t_i) = 
    \begin{pmatrix}
        \cos \theta_i^{t} & -\sin\theta_i^{t} \\
        \sin\theta_i^{t} & \cos \theta_i^{t}
    \end{pmatrix}.
\end{equation}
While RoPE effectively preserves the relative positional information by maintaining dot-product structure between queries and keys, it introduces complexity in quantization due to the mixing of paired channel distributions. This can cause instability, especially in low-bit quantization scenarios with pronounced outliers.



\subsection{Hardware Background: Hopper GPU and FlashAttention-3}
NVIDIA’s Hopper GPU (H100) features FP8-compatible tensor cores optimized for high-throughput, low-precision matrix multiplication, facilitating efficient inference of quantized LLMs. These cores natively support FP8 formats and achieve double throughput for FP16/BF16 operations, which is particularly beneficial during epilogue computations. FlashAttention-3~\citep{shah2024flashattention} is a memory-efficient attention kernel designed specifically for FP8 arithmetic. It employs a fused softmax and load-store operation organized into a two-stage pipeline to maximize performance.
In our work, we extend FlashAttention-3 by introducing a three-stage pipeline for the attention kernel, enhancing memory utilization and compute efficiency. Additionally, we generalize kernel support to mixed-precision INT4 $\times$ FP8 GEMM operations, further optimizing performance for both linear layers and the prefill phase of the attention mechanism.



%% file: sections/03-method.tex
\begin{figure}[t]
    \centering
    \includegraphics[width=1.0\linewidth]{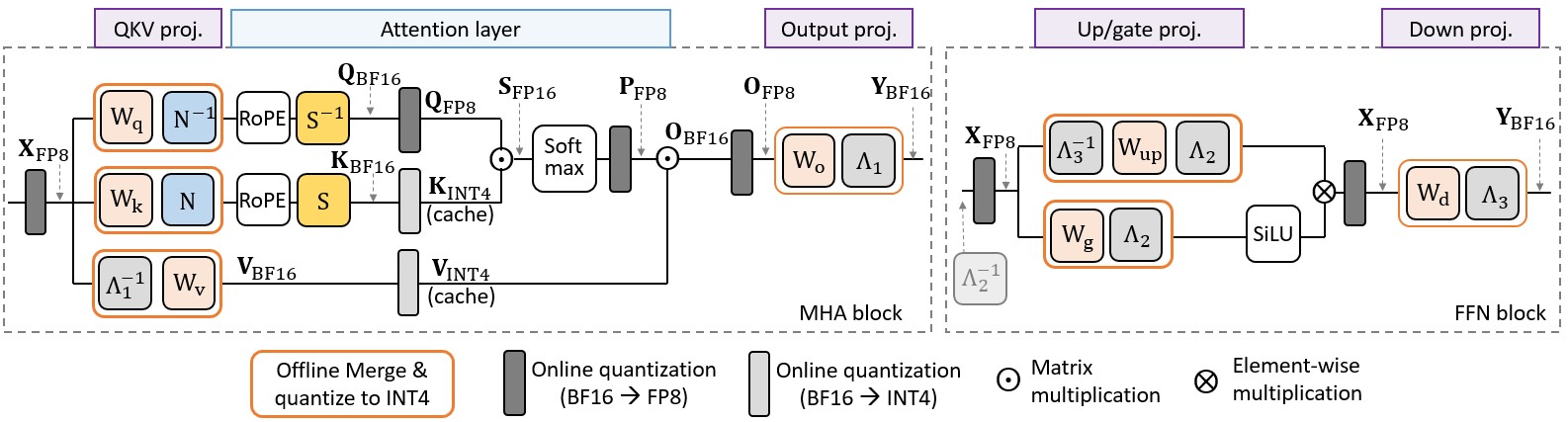}
    \vspace{-0.3cm}
    \caption{\small Overall diagram of mixed-precision quantization and corresponding outlier smoothing strategies. $\Lambda$, N, and S denote channel-wise absmean scaling (CAS), RoPE-preserving normalization (RPN), and Channel-wise RoPE scaling (CRS), respectively. All weights are merged with calibration matrices and quantized offline as shown in orange boxes, incurring no runtime overhead.} 
    \label{fig:overall_quantization}
\end{figure}


\section{Method}\label{sec:method}

In this section, we present the detailed design of \sysname, a co-designed framework combining INT4-FP8 quantization and kernel-level optimizations to accelerate LLM inference. We first introduce our optimized INT4-FP8 GEMM kernel and quantization strategy. Then, we describe the three-stage pipelined kernel for the prefill phase and RoPE-aware quantization strategy for attention layers.

\subsection{Linear Layer}
\sysname adopts a co-designed mixed-precision approach, quantizing linear layer weights to INT4 and activations to FP8. Specifically, weights are quantized using per-token group-wise symmetric quantization with groups of 128 elements sharing a single FP8  (E4M3)  scaling factor. Activations are quantized in FP8 (E4M3) format using a BF16 scaling factor. Section~\ref{subsubsec:int4fp8} details the corresponding GEMM kernel implementation. 

\subsubsection{Kernel optimization: INT4 $\times$ FP8.} \label{subsubsec:int4fp8}

\begin{figure}[h]
    \centering
    \includegraphics[width=0.75\linewidth]{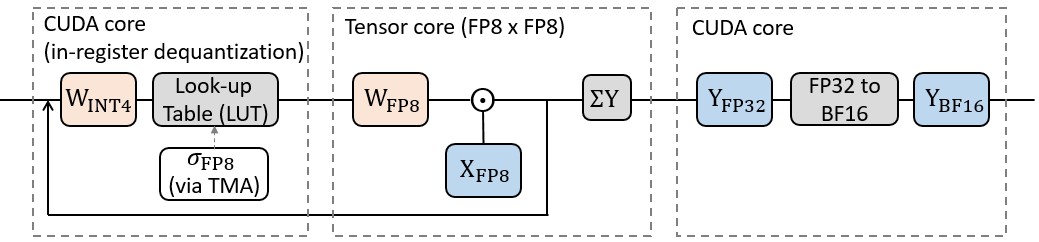}
    \caption{\small INT4 $\times$ FP8 GEMM kernel for linear layer computations.} 
    \label{fig:gemm_kernel}
\end{figure}

Figure~\ref{fig:gemm_kernel} illustrates the INT4 $\times$ FP8 GEMM kernel structure, which is applied across all linear layers (e.g., up, down, gate projections, QKV generation, and output projection) as well as for attention score ($\mathbf{S}=\mathbf{QK^{\text {T}}}$) and output ($\mathbf{O}=\mathbf{PV}$) computations (Figure~\ref{fig:overall_quantization}).

We implement this optimized kernel using CUTLASS library through three steps (Figure~\ref{fig:gemm_kernel}): 
\begin{compactenum}[$\circ$]
    \item Step 1: A lookup table (LUT) of 16 FP8 entries is constructed based on the group-wise FP8 scaling factor $\sigma$, mapping INT4 values $\{-8\sigma, -7\sigma, ..., 6\sigma, 7\sigma\}.$ INT4 weights and corresponding FP8 scaling factors are asynchronously loaded into shared memory using the Tensor Memory Accelerator (TMA), and INT4 weights are converted to FP8 in registers by CUDA cores.
    \item Step 2: The tensor cores perform GEMM by multiplying FP8 input activations with FP8-converted weights, accumulating intermediate partial sums in FP32 to minimize precision loss.
    \item Step 3: CUDA cores reduce GEMM outputs from FP32 to BF16 for efficient epilogue computation. This includes activation addition, nonlinear activation functions (e.g., ReLU, SiLU), and element-wise multiplications. Leveraging Hopper GPUs’ optimized FP16/BF16 computation intrinsics and assemblies, this step maximizes throughput.
\end{compactenum}

The selection of $(\sigma_{\text{FP8}})$ is a key contribution of \sysname for accelerating inference. A detailed experimental analysis is provided in Section~\ref{subsec:FFNperformance}, and the rationale is discussed in Appendix~\ref{ap:rationaleforscalingfactor}.

\subsubsection{Linear layer scaling strategy}\label{subsubsec:linearlayerscaling}

Although the FP8 scaling factor $\sigma_{\text{FP8}}$ improves throughput in quantized model serving, it may introduce underflow and overflow issues, as described in Appendix~\ref{ap:underflowandoverflowbyfp8}. To address underflow caused by FP8 scaling factors ($\sigma_{\text{FP8}}$) and enhance INT4 quantization accuracy, we propose two scaling strategies: \textbf{channel-wise absmean scaling (CAS)} and \textbf{per-tensor scaling (PTS).} \textbf{Channel-wise absmean scaling} normalizes each channel's distribution to a shared target absolute mean, reducing inter-channel variance and improving group quantization stability.

\smallskip
\begin{defi}[\texttt{Channel-wise Absmean Scaling, CAS}]\label{def:channel-wise-absmean-scaling}
   Let $\mathbf{W}$ be a weight matrix, and $\mathbf{\Lambda}$ a diagonal scaling matrix. The scaled weight matrix is given by:
    \begin{equation}
    \bar{\mathbf{W}} = \mathbf{W}\mathbf{\Lambda}, \quad \text{and} \quad \lambda^i \triangleq \frac{\bar{\omega}}{\text{absmean}(\omega^i)},
   \end{equation}   
   where $\lambda^i$ is the $i$-th diagonal element of $\Lambda$, ${\omega}^i$ is the $i$-th channel of $\mathbf{W}$, and $\bar{\omega}$ the target absmean.
\end{defi}  
To maintain computational equivalence, we apply the inverse scaling to activations:
\begin{equation}
    \mathbf{Y} = (\mathbf{X\Lambda^{-1}})_{\text{FP8}}\cdot(\mathbf\Lambda^{\text{T}}\mathbf{W}^{\text{T}})_{\text{INT4}} =\mathbf{X}\mathbf{W}^{\text{T}},
\end{equation}
where $(\cdot)_{\text{precision}}$ indicates the representation precision of the tensor. As illustrated in Figure~\ref{fig:overall_quantization}, similar to how $\Lambda$ is merged offline with weights prior to quantization, the inverse scaling factor $\Lambda^{-1}$ can also be integrated offline with the preceding layer's weights, enabling offline merging and quantization. In Figure~\ref{fig:overall_quantization}, for $\Lambda_1$, $\lambda^i$ is exceptionally set as a constant for quantization stability.

To further mitigate underflow risks, we introduce \textbf{per-tensor scaling}, globally adjusting the scale of weights. We choose the smallest scaling factor $\delta = 2^n$ to balance underflow prevention and overflow risk as follows:


\smallskip
\begin{defi}[\texttt{Per-tensor Scaling for Preventing Underflow, PTS}]\label{def:per-tensor-scaling}
Let $\mathbf{W}$ be a weight matrix. We define $n$ as the smallest non-negative integer meeting one of the following conditions:
\begin{compactenum}[1.]
\item For all $i \in \mathbb{N},$
\begin{equation}\label{eq:pts_first_condition}
    \mathbb{S}(\mathbf{W}\cdot 2^n) = \mathbb{S}(\mathbf{W}\cdot 2^{n+i}), \quad \text{and} \quad \mathbb{S}(\mathbf{W}) \triangleq \sum_{\omega \in \mathbf{W}} \max\left( 0,\; 7\cdot 2^{-9} - |\omega|\right)
\end{equation}
measures cumulative underflow relative to FP8's minimum representable value.
\smallskip
\item There exists an element $\omega \in \mathbf{W}$ satisfying:
\begin{equation}\label{eq:pts_second_condition}
7 \cdot 2^{5-n} \leq |\omega| < 7 \cdot 2^{6-n},
\end{equation}
indicating overflow risk at scaling beyond $\delta =2^n$.
\end{compactenum}
\end{defi}
Per-tensor scaling ensures that as many weight elements $\omega$ as possible fall within the representable range of FP8. The weight matrix $\mathbf W$ is scaled by $\delta$ before quantization, and the inverse scaling $\delta^{-1}$ is simply multiplied to the GEMM output. A detailed derivation is provided in Appendix~\ref{appedix:per-tensor-scaling}.

\subsection{Attention Layer}\label{subsection:attentionlayerrope}

\subsubsection{RoPE-aware quantization and outlier smoothing}\label{subsubsec:rope_aware_quantization}

\begin{figure}[h]
\centering
    \subfloat[\small Pre-RoPE Q.]{\includegraphics[width=0.17\columnwidth]{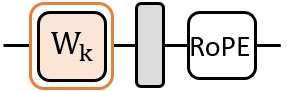}\label{fig:pre_rope_quantization}}
    \hfill
    \subfloat[\small Post-RoPE Q.]{\includegraphics[width=0.16\columnwidth]{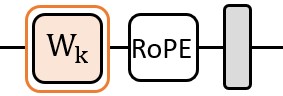}\label{fig:post_rope_quantization}}
    \hfill
    \subfloat[RPN]{\includegraphics[width=0.19\columnwidth]{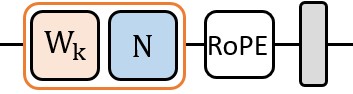}\label{fig:post_rope_quant_pre_rope_normalization}}
    \hfill
    \subfloat[CRS]{\includegraphics[width=0.21\columnwidth]{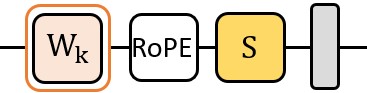}\label{fig:post_rope_quant_post_rope_normalization}}
    \hfill
    \subfloat[RPN + CRS (\sysname)]{\includegraphics[width=0.235\columnwidth]{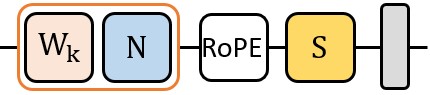}\label{fig:post_rope_quant_pre_post_rope_normalization}}
\caption{RoPE-aware quantization and outlier smoothing strategies for key matrix.}
\label{fig:rope-aware-quantization}
\vspace{-0.2cm}
\end{figure}

To efficiently quantize attention layers, \sysname quantizes query vectors to FP8 and key-value pairs to INT4. However, quantizing key-value matrices poses unique challenges due to the rotary positional embeddings, which significantly alter the statistical distributions of query and key vectors.  Figure~\ref{fig:rope-aware-quantization} illustrates various RoPE-aware quantization strategies for the key matrices.

\noindent \textbf{Post-RoPE quantization.}
\sysname adopts post-RoPE quantization (Figure~\ref{fig:post_rope_quantization}) as a baseline, applying per-token (row-wise) quantization where each row has its scaling factor $\sigma$. In contrast, pre-RoPE quantization (Figure~\ref{fig:pre_rope_quantization}) requires additional quantization and dequantization steps, because the quantized key cache must be dequantized before applying RoPE, and then re-quantized afterward. Post-RoPE quantization reduces overhead by eliminating these extra operations. Nonetheless, designing effective outlier smoothing under post-RoPE quantization is not straightforward. To address this, we propose a two-stage strategy (Figure~\ref{fig:post_rope_quant_pre_post_rope_normalization}) that combines both pre- and post-RoPE smoothing techniques.

\noindent \textbf{Two-stage RoPE-aware outlier smoothing.}
As RoPE intermixes paired channel distributions while leaving non-paired channels unaffected, we separate key matrix channels into two categories: (i) outlier channels with their paired counterparts, and (ii) regular channel pairs. We apply Channel-wise RoPE Scaling (CRS) post-RoPE to the first group, and RoPE-preserving normalization (RPN) pre-RoPE to the second group.

\smallskip
\noindent \textit{Stage 1. RoPE-preserving normalization (RPN):} For each RoPE channel pair $(\mathbf{k}_i, \mathbf{k}_j)$ within a head, RPN calculates the maximum $L_2$ norm across tokens and applies a common scaling factor ensuring a bounded norm of post-RoPE (Theorem~\ref{thm:rope_preserving_normalization}). As a linear, offline-calibrated transformation, RPN can be fused into model weights without runtime overhead (Figure~\ref{fig:post_rope_quant_pre_rope_normalization} and Figure~\ref{fig:post_rope_quant_pre_post_rope_normalization}), improving dynamic range utilization and reducing INT4 quantization clipping risks.

\begin{theorem}[\texttt{RoPE-preserving Normalization Bound}] \label{thm:rope_preserving_normalization}
Given a key matrix $\mathbf{K} \in \mathbb{R}^{N \times d}$, where $N$ is the sequence length and $d$ is the head dimension, consider a RoPE channel pair $(\mathbf{k}_i, \mathbf{k}_j) \in \mathbb{R}^{N \times 2}$ with $j = i + \frac{d}{2}$.  
Then, for $\alpha > 0$, the scaled key channels satisfy:
\begin{equation}
\left\| \left( \frac{k_i^n}{s_i}, \frac{k_j^n}{s_j} \right) \right\|_2 \leq \frac{1}{\alpha}, ~\forall n \in \{1,\dots N\}, \quad \text{and} \quad s_j = s_i \triangleq \alpha \cdot \max_{n \in \{1, \dots, N\}} \left\| \left(k_i^n, k_j^n\right) \right\|_2, 
\end{equation}
where $s_i$ (or $s_j$) is the shared scaling factor of channel $i$ and $j.$ That is, RPN ensures that scaled channel pairs have their norms bounded within a radius of ${1}/{\alpha}.$
\end{theorem}

Theorem~\ref{thm:rope_preserving_normalization} highlights a distinct advantage of our approach: the uniform boundedness of channel norms of post-RoPE, which reduces inter-channel variance and effectively smooths outliers. A simple proof of this theorem is provided in Appendix~\ref{ap:proofoftheorem}.

\noindent \textit{Stage 2. Channel-wise RoPE Scaling (CRS):} CRS applies distinct scaling factors independently to outlier channels and their pairs post-RoPE, effectively smoothing outliers. For outlier channel $\mathbf{k}_i = \{k_i^n\}_{n\in \{1, \dots, N\}},$ the CRS factor $t_i$ is defined by: $t_i \triangleq \beta \cdot \max_{n \in \{1, \dots, N\}}|k_i^n|$ for $\beta >0.$ Unlike RPN, CRS incurs slight online computational overhead (Figure~\ref{fig:post_rope_quant_post_rope_normalization} and Figure~\ref{fig:post_rope_quant_pre_post_rope_normalization}), but its limited application scope ensures that the overhead remains negligible.

We note that for computational equivalence, the inverse calibration matrices, RPN ($N^{-1}$) and CRS ($S^{-1}$), are merged in the query projection layer, as shown in Figure~\ref{fig:overall_quantization}.

\subsubsection{Modified FlashAttention-3: prefill phase}

To further optimize inference performance, \sysname extends the original two-stage pipeline of FlashAttention-3~\citep{shah2024flashattention} to a three-stage pipelined approach, as depicted in Figure~\ref{fig:3-stage-pipelining}. The overall procedure of prefill phase consists of two main warpgroup components: the \texttt{Producer} and the \texttt{Consumer}, each responsible for distinct stages of data processing.

\noindent \textbf{Producer warpgroup (data preparation).} The producer warpgroup asynchronously loads the Q, K, and V matrices from global memory (\texttt{GMEM}) to shared memory (\texttt{SMEM}), using the TMA. The value matrix is transposed during loading to optimize subsequent GEMM operations, ensuring minimal latency and efficient data flow for the consumer warpgroup. 

\noindent \textbf{Consumer warpgroup (three-stage computation).} The consumer warpgroups (Algorithm~\ref{al:3-phase-fmha}) execute computations asynchronously in three overlapping stages (Figure~\ref{fig:3-stage-pipelining}):

\begin{figure}[t]
    \centering
    \includegraphics[width=0.9\linewidth]{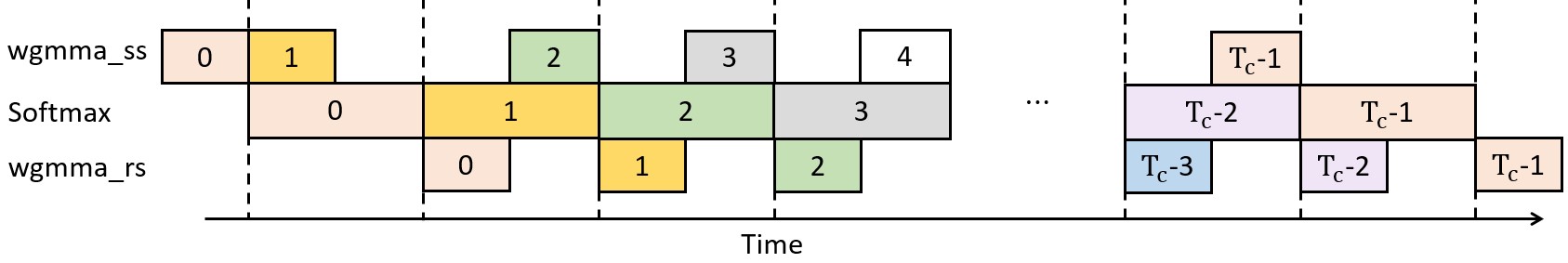}
    \caption{\small 3-stage \texttt{wgmma}\_\texttt{ss}-Softmax-\texttt{wgmma}\_\texttt{rs} pipelining.} 
    \label{fig:3-stage-pipelining}
\end{figure}

\begin{compactenum}[$\circ$]
    \item \text{Stage 1. Query-Key multiplication $(\textbf{S}=\textbf{QK}^{\text{T}})$ (lines 3 to 8):} GEMM operations are performed directly from \texttt{SMEM} using $\texttt{wgmma}\_\texttt{ss}$, with results accumulated in FP16. Computed attention scores serve as inputs to Stage 2 without waiting, enabling pipelined execution.
    \item \text{Stage 2. Softmax computation $(\textbf{P}_{ij} = \text{softmax}(\textbf{S}_{ij}))$ (lines 9 to 17):} Attention scores from Stage 1 are normalized via softmax in FP16 and subsequently quantized to FP8 for efficient storage in registers. Meanwhile, the subsequent query-key multiplication begins concurrently in Stage 1.
    \item \text{Stage 3. Value aggregation $(\textbf{O} =\textbf{PV})$ (lines 18 to 24):} The FP8 softmax matrix $(\textbf{P}_{ij})$ multiplies the transposed value matrix $\text{V}^{\text {T}}$ using $\texttt{wgmma}\_{\texttt{rs}}$, involving register-to-\texttt{SMEM} operations. Intermediate results accumulate in FP32, then reduce to BF16 for final storage back into \texttt{GMEM}.
\end{compactenum}

Each computational stage is executed asynchronously and synchronized by explicit \texttt{wait} instructions to ensure correct data dependencies. To optimize memory usage and throughput further, we reduce the precision of the accumulated $\mathbf{S}_{ij}$ values and the row-wise maximum values $m_i$ from FP32 to FP16, utilizing FP16 exponential instructions. Although this precision reduction might introduce minimal accuracy loss, the subsequent FP8 quantization of softmax results $\mathbf{P}_{ij}$ ensures sufficient accuracy for the final GEMM step. We retain FP32 precision only for critical intermediate accumulations ($s_i$, $l_i$, and $\mathbf{O}_i$) to preserve accuracy in the final result. 
We note that a complete algorithm, including the behavior of the \texttt{producer warpgroup}, is provided in Appendix~\ref{ap:full_prefill_phase}.

\begin{algorithm}[t]
\caption{Three-stage attention layer forward at prefill phase in \texttt{consumer warpgroup}}
\label{al:3-phase-fmha}
\small
\begin{spacing}{1.15}
\begin{algorithmic}[1]
\Require:$\mathbf{Q}_i\in \mathbb{R}^{B_r \times d}, \mathbf{K}, \mathbf{V} \in \mathbb{R}^{N \times d}$, query block size $B_r$, key and value block size $B_c$ with $T_c=\lceil N/B_c \rceil$
\State Initialize pipeline objects to manage barrier synchronization with a 2-stage \texttt{SMEM} buffer
\State Fetch the query index $i$ from global memory with atomic decrement
\State Allocate the predetermined number of registers
\State Initialize $\mathbf{O}_{ik}=(0)\in \mathbb{R}^{B_r\times d}$ for $k \in \{0,1\}$, 
~$s_i,l_i,m_i=(1),(0),(-\infty)\in\mathbb{R}^{B_r}$
\State Commit $\textbf{S}_{i0}=\texttt{wgmma}\_\texttt{ss}(\mathbf{Q}_{ik}, \mathbf{K}_0)$ and \texttt{wait} $\textbf{S}_{i0}$
\State Commit $\textbf{S}_{i1}=\texttt{wgmma}\_\texttt{ss}(\mathbf{Q}_{ik}, \mathbf{K}_1)$ and do not \texttt{wait} $\textbf{S}_{i1}$
\State Compute $\textbf{S}_{i0}=\text{mask}(\tau \textbf{S}_{i0})$, $m_i=\max\left(\text{rowmax}(S_{i0})\right), ~\textbf{P}_{i0}=\exp(\textbf{S}_{i0}-m_i),~l_i=\text{rowsum}(\textbf{P}_{i0})$
    \State \texttt{Wait} $\textbf{S}_{i1}$
\For{$2\leq j < T_c$}
    \State Commit $\mathbf{O}_{ik}=\mathbf{O}_{ik}+\texttt{wgmma}\_\text{rs}\left(\textbf{P}_{i(j-2)}, \mathbf{V}_{j-2}^\text{T}\right)$ and do not \texttt{wait} $\mathbf{O}_{ik}$
    \State Compute $\textbf{S}_{i(j-1)}=\text{mask}(\tau \textbf{S}_{i(j-1)})$
    \State $m_{\text{old}}=m_i$ and compute $m_i=\max\left(m^{\text{old}}_i, \text{rowmax}(\textbf{S}_{i(j-1)})\right),~s_i=\exp(m^{\text{old}}_i-m_i), ~l_i=s_i\odot l_i$
    \State Commit $\textbf{S}_{ij}=\texttt{wgmma}\_\texttt{ss}(\textbf{Q}_{ik}, \mathbf{K}_j)$ and do not \texttt{wait} $\textbf{S}_{ij}$
    \State Compute $\textbf{P}_{i(j-1)}=\exp(\textbf{S}_{i(j-1)}-m_i),~l_i=l_i+\text{rowsum}(\textbf{P}_{i(j-1)})$
    \State \texttt{Wait} $\mathbf{O}_{ik}$ and compute $\mathbf{O}_{ik}=s_i\odot \mathbf{O}_{ik}$
    \State \texttt{Wait} $\textbf{S}_{ij}$
\EndFor
\State Commit $\mathbf{O}_{ik}=\mathbf{O}_{ik}+\texttt{wgmma}\_\texttt{rs}\left(\textbf{P}_{i(T_c-2)}, \mathbf{V}_{T_c-2}^\text{T}\right)$ and do not \texttt{wait} $\mathbf{O}_{ik}$
\State $m^{\text{old}}_i=m_i$ and compute $m_i=\max\left(m^{\text{old}}_i, \text{rowmax}(\textbf{S}_{i(T_c-1)})\right),~s_i=\exp(m^{\text{old}}_i-m_i), ~l_i=s_i\odot l_i$
\State Compute $\textbf{P}_{i(j-1)}=\exp(\textbf{S}_{i(T_c-1)}-m_i),~l_i=l_i+\text{rowsum}(\textbf{P}_{i(T_c-1)})$
\State \texttt{Wait} $\mathbf{O}_{ik}$ and compute $\mathbf{O}_{ik}=\mathbf{s}_i\odot \mathbf{O}_{ik}$
\State Commit $\mathbf{O}_{ik}=\mathbf{O}_{ik}+\texttt{wgmma}\_\texttt{rs}\left(\mathbf{P}_{i(T_c-1)}, \mathbf{V}_{T_c-1}^\text{T}\right)$ and \texttt{wait} $\mathbf{O}_{ik}$
\State Compute $\mathbf{O}_{ik}=1/l_i \odot \mathbf{O}_{ik}$
\State TMA store $\mathbf{O}_{ik}$ from \texttt{SMEM} to \texttt{GMEM}
\end{algorithmic}
\end{spacing}
\end{algorithm}


%% file: sections/04-experiments.tex
\section{Evaluation}\label{sec:evaluation}

\subsection{Evaluation Setup}

We evaluate \sysname on a range of LLMs, including Llama3.1 (8B), Llama3 (8B)~\citep{llama3}, Llama2 (7B, 13B)~\citep{touvron2023llama2}. All experiments are conducted on NVIDIA H100 SXM5 GPUs, configured with CUDA 12.6 and PyTorch 2.6.0 on a 2-socket Intel(R) Xeon(R) Platinum 8468 CPUs with 2TB DDR5 memory system by PCIe Gen5 X16 links. All of the experiments are conducted on a single GPU, without using PCIe or NVLink for tensor inter-GPU communications. Kernel optimization for linear layers and prefill phase employs INT4 $\times$ FP8 GEMM operations implemented using the CUTLASS library~\citep{cutlass}. To assess the accuracy of \sysname, we employ the WiKiText-2~\citep{merity2016pointerwikitext} dataset to measure perplexity (PPL), while for downstream tasks, we adopt six different zero-shot evaluation including PIQA~\citep{bisk2020piqa}, ARC-Easy~\citep{clark2018thinkarc}, ARC-Challenge~\citep{clark2018thinkarc}, BoolQ~\citep{clark2019boolq}, HellaSwag~\citep{zellers2019hellaswag}, and WinoGrande~\citep{sakaguchi2021winogrande}. 
Throughput is evaluated on the FFN layer and the prefill phase, by measuring the number of generated tokens per second, averaged over ten repeated runs. For accuracy and throughput comparisons, we include state-of-the-art quantization methods: INT4-FP8 AWQ~\citep{lin2024awq} implemented with TensorRT-LLM v0.18.2, QServe~\citep{lin2024qserve}, Atom~\citep{zhao2024atom}, and QuaRot~\citep{ashkboos2024quarot}.

\begin{figure}[t]
    \centering
    \includegraphics[width=1.0\linewidth]{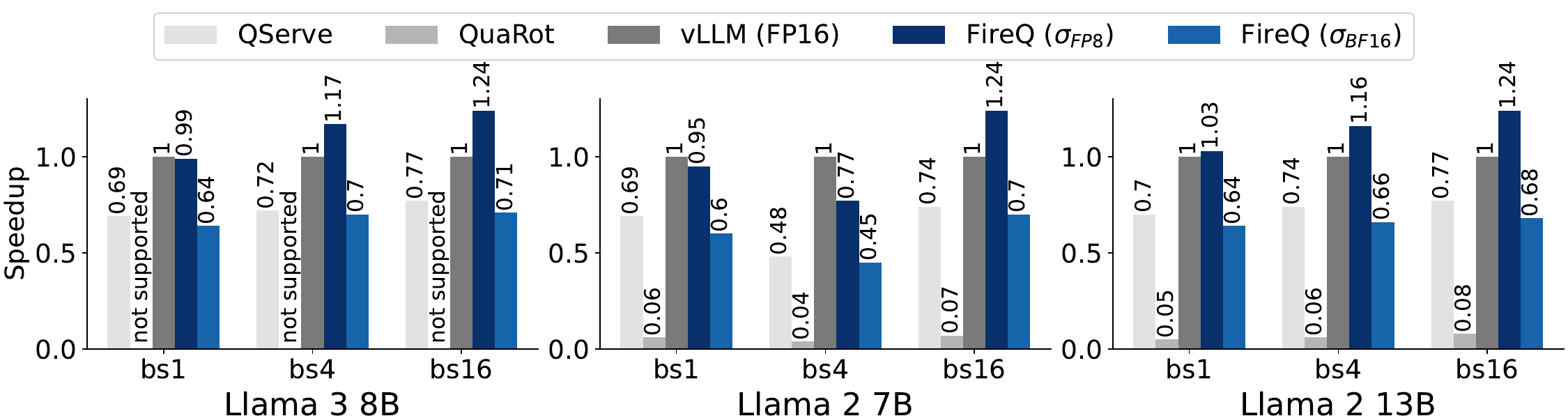}
    \caption{Impact of INT4-FP8 GEMM kernel on feed-forward network layers. The x-axis represents the batch size, with throughput results in the y-axis normalized to the FP16 baseline performance.} 
    \label{fig:int4-fp8-kernel}
\end{figure}

\subsection{Impact of INT4-FP8 Kernel}\label{subsec:FFNperformance}

We investigate the performance improvements provided by \sysname’s INT4-FP8 GEMM kernel by evaluating throughput in feed-forward network layers. As a baseline, we use non-quantized FP16 models evaluated with the vLLM framework~\citep{vllm}. Typically, quantized models exhibit reduced throughput compared to non-quantized counterparts due to quantization and dequantization overheads on CUDA cores, especially in GEMM operations. However, as illustrated in Figure~\ref{fig:int4-fp8-kernel}, \sysname in most cases surpasses the throughput of the non-quantized FP16 baseline. This occurs because \sysname not only minimizes performance penalties associated with dequantization but also benefits from reduced memory accesses enabled by lower-bit quantization. Furthermore, when compared with the state-of-the-art framework QServe~\citep{lin2024qserve}, \sysname achieves a 1.68× throughput improvement at the Llama2-7B batch size 16 case. This performance enhancement underscores the effectiveness of our INT4-FP8 GEMM kernel, which employs FP8 scaling factors ($\sigma_{\text{FP8}}$) for INT4 quantization, enabling efficient in-register dequantization with minimal overhead, thus outperforming QServe’s W4A8-INT operations. Additionally, we include a comparison using BF16 scaling factors ($\sigma_{\text{BF16}}$) to clearly illustrate the throughput advantage derived specifically from our FP8 scaling approach. A detailed discussion of $\sigma_{\text{BF16}}$ is provided in Appendix~\ref{ap:discussionBF16scalingfactor}.

\subsection{Impact of Prefill Phase Optimization}\label{subsec:impact_of_prefill_phase_optimization}

We evaluate the performance of \sysname's prefill phase kernel, emphasizing improvements in inference throughput. Our benchmarks include the Llama3-8B model tests at a sequence length of 2048 tokens. We compare \sysname against a non-quantized FP16 baseline (TensorRT-LLM FP16) and QServe~\citep{lin2024qserve}. Given the memory-intensive nature of prefill operations, quantized models generally offer better throughput by reducing memory latency than non-quantized models. Our results demonstrate that \sysname outperforms the comparison frameworks. Specifically, at a sequence length of 1024, \sysname achieves throughput improvements of up to 1.29x over the FP16 baseline and 1.26x over QServe, as described in Table~\ref{tab:prefill_phase_throughput}. Moreover, at longer sequence lengths (i.e., 2048, 4096), FireQ significantly accelerates first-token generation, achieving 2.53× and 2.38× speedups, respectively, compared to the original FP16 model. Also, the scaled-dot product kernel performance is provided in Appendix~\ref{ap:scaled-dot-product}.
These improvements primarily arise from our effective utilization of tensor cores for attention computations, contrasting with other frameworks relying on CUDA cores. 
\begin{table}[h]
\centering
\caption{\small Prefill phase throughput, Llama3-8B, batch size = 16}
\label{tab:prefill_phase_throughput}
\resizebox{0.8\textwidth}{!}{%
\renewcommand{\arraystretch}{1.2}
\begin{tabular}{ccccc}
\toprule
\textbf{} & \text{Seq. length} & \text{TRT-LLM (FP16)} & \text{QServe} & \textbf{\sysname} \\
\midrule
\multirow{3}{*}{Tokens/sec (speedup)} & 1024 & 28.8k (baseline) & 29.4k (1.02x) & 37.2k \textbf{(1.29x)} \\
 & 2048 & 13.6k (baseline) & 29.4k (2.16x) & 34.5k \textbf{(2.53x)} \\
 & 4096 & 13.8k (baseline) & 28.5k (2.06x) & 32.9k \textbf{(2.38x)} \\
\bottomrule
\end{tabular}%
}
\end{table}

\begin{table}[t]
\centering
\caption{Zero-shot accuracy on six common sense tasks for Llama3-8B}
\label{tab:zero-shot-accuracy}
\resizebox{0.9\textwidth}{!}{%
\renewcommand{\arraystretch}{1.2}
\begin{tabular}{cccccccccc}
\toprule
\textbf{Precision} & \textbf{Algorithm} & \textbf{PPL $\downarrow$} & \textbf{PIQA}$\uparrow$ & \textbf{ARC-e}$\uparrow$ & \textbf{ARC-c}$\uparrow$ & \textbf{BoolQ}$\uparrow$ & \textbf{HS}$\uparrow$ & \textbf{WG}$\uparrow$ & \textbf{Avg.}$\uparrow$ \\
\midrule
 FP16 & FP16 & 6.13 & 79.54 & 80.09 & 50.17 & 81.35 & 60.18 & 72.61 & 70.61 \\
\midrule
W4A4 & Atom & 7.57 & 76.33 & 77.65 & 44.28 & 76.45 & 55.22 & 69.37 & 66.55\\
\midrule
W4A8 & QServe & 6.70 & 77.80 & 77.78 & \textbf{48.12} & \textbf{80.49} & \textbf{58.73} & 71.59 & \textbf{69.08} \\
\midrule
\multirow{3}{*}{W4A8-FP}& TRT-LLM INT4-FP8 & - & 77.15 & 76.64 & 45.73 & 73.21 & 58.44 & 72.45 & 67.25 \\
 & \cellcolor{gray!10} \sysname($\sigma_{\text{FP8}}$) & \cellcolor{gray!10} 7.06 & \cellcolor{gray!10} 77.74 & \cellcolor{gray!10} \textbf{78.15} & \cellcolor{gray!10} 45.64 & \cellcolor{gray!10} 75.07 & \cellcolor{gray!10} 58.10 & \cellcolor{gray!10} \textbf{72.77} & \cellcolor{gray!10} 67.92 \\
 & \cellcolor{gray!10} \sysname($\sigma_{\text{BF16}}$) & \cellcolor{gray!10} 7.03 & \cellcolor{gray!10} \textbf{78.62} & \cellcolor{gray!10} 76.98 & \cellcolor{gray!10} 45.14 & \cellcolor{gray!10} 75.44 & \cellcolor{gray!10} 58.35 & \cellcolor{gray!10} 71.98 & \cellcolor{gray!10} 67.75 \\
\bottomrule
\end{tabular}%
}
\end{table}

\subsection{Accuracy Evaluation}

We evaluate the impact of \sysname on model accuracy to demonstrate that the high-throughput quantization design of \sysname does not compromise inference quality. We benchmark the accuracy of \sysname against both the FP16 non-quantized baseline and several state-of-the-art quantization frameworks, including QServe, Atom, and TensorRT-LLM’s AWQ-based INT4-FP8 models. Our evaluation across standard zero-shot tasks, including PIQA, ARC-e, ARC-c, BoolQ, HellaSwag (HS), and WinoGrande (WG), reveals that \sysname consistently maintains accuracy comparable to other frameworks. Specifically, \sysname achieves minimal accuracy degradation while significantly outperforming other quantized methods in throughput. Notably, \sysname obtains higher average scores on zero-shot benchmarks compared to TensorRT-LLM’s INT4-FP8 framework, and achieves the highest individual scores on PIQA, ARC-e, and WinoGrande benchmarks among all evaluated methods. The perplexity (PPL) of \sysname is approximately 0.3 higher than QServe, but this trade-off is justified by \sysname’s substantial throughput improvements. These results highlight the effectiveness of our carefully designed scaling strategies and RoPE-aware quantization, emphasizing \sysname’s ability to deliver superior inference performance without meaningful loss in model accuracy.

\subsection{Ablation Study for Quantization Strategies}

To thoroughly analyze the impact of our quantization strategies on model accuracy, we conduct an ablation study using the Llama3.1-8B model evaluated on the WikiText2 dataset (Table~\ref{tab:ablation}). We systemically address the impact of various outlier smoothing and scaling techniques in \sysname, including RoPE-preserving normalization (RPN), channel-wise RoPE scaling (CRS), channel-wise absmean scaling (CAS), and per-tensor scaling (PTS). Table~\ref{tab:ablation} summarizes the perplexity scores. Individual applications of RPN, CRS, CAS, and PTS each reduce perplexity to varying degrees, highlighting their standalone effectiveness. Combining these strategies yields further improvements, with the comprehensive integration of RPN, CRS, CAS, and PTS achieving the lowest perplexity. The further detailed analysis for these strategies is provided in Appendix~\ref{ap:ablation_for_smoothing_strategies}. Additionally, we compare the standard FP8-based \sysname with variants using BF16 scaling factors ($\sigma_{\text{BF16}}$) and BF16 tensor cores. Using BF16 scaling factors alone marginally improves perplexity over FP8 scaling factors, but introduces a trade-off in inference throughput, as discussed in Figure~\ref{fig:int4-fp8-kernel}.

\begin{table}[h]
\centering
\caption{WikiText2 perplexity of Llama3.1-8B}
\label{tab:ablation}
\resizebox{1.0\textwidth}{!}{%
\renewcommand{\arraystretch}{1.2}
\begin{tabular}{ccccccccc}
\toprule
\textbf{\sysname variants}& \textbf{No smoothing} & \textbf{RPN} & \textbf{CRS} & \textbf{CAS} & \textbf{PTS} & \textbf{RPN+CRS} & \textbf{RPN+CRS+CAS} & \textbf{RPN+CRS+CAS+PTS} \\
\midrule
 \sysname ($\sigma_{\text{FP8}}$)& 8.15& 8.00&7.48 &7.68 &7.98 &7.44 &7.18 &7.11 \\
 \sysname ($\sigma_{\text{BF16}}$)&7.87 &7.69 &7.41 & 7.50& 7.69 &7.34 & 7.09&7.09 \\
\bottomrule
\end{tabular}%
}
\end{table}


%% file: sections/05-related_work.tex
\section{Related Work}

Post-training quantization enhances inference performance for LLMs. SmoothQuant~\citep{xiao2023smoothquant} introduces a \text{W8A8} scheme flattening activations during weight quantization, while AWQ~\citep{lin2024awq} proposes a \text{W4A16} strategy quantizing activation-aware weights with FP16 activations. QuaRot~\citep{ashkboos2024quarot}, SpinQuant~\citep{liu2024spinquant}, and Atom~\citep{zhao2024atom} implement \text{W4A4} quantization using Hadamard transformations to manage outliers in linear and attention layers. Other approaches employ rotations to mitigate outliers in activations (QServe~\citep{lin2024qserve}, \text{W4A8KV4}) or KV caches (KVQuant~\citep{hooper2024kvquant}, RotateKV~\citep{su2025rotatekv}).

RoPE-aware quantization and KV scaling are extensively explored in frameworks like QServe~\citep{lin2024qserve}, KVQuant~\citep{hooper2024kvquant}, RotateKV~\citep{su2025rotatekv}, QuaRot~\citep{ashkboos2024quarot}, and SpinQuant~\citep{liu2024spinquant}. \sysname combines post-RoPE quantization and pre-RoPE calibration similar to QServe, QuaRot, and SpinQuant for enhanced throughput, and post-RoPE calibration from QServe, KVQuant, and RotateKV for accuracy. It specifically applies channel-wise scaling factors that consider RoPE effects during KV cache calibration.

While previous methods primarily use integer quantization due to constraints of the Ampere architecture, \sysname leverages the Hopper architecture’s FP8 tensor cores by implementing FP8 activations with optimized INT4 $\times$ FP8 kernels. \sysname applies FP8 scaling factors to INT4 quantization, accelerating GEMM operations on FP8 tensor cores. Additionally, it optimizes INT4 $\times$ FP8 kernels using the CUTLASS library and extends FlashAttention-3~\citep{shah2024flashattention} by upgrading its two-stage pipeline to a three-stage pipeline, significantly enhancing prefill inference throughput.

%% file: sections/06-conclusions.tex
\section{Conclusions}\label{sec:conclusion}

In this paper, we proposed \sysname, an efficient INT4-FP8 quantization and kernel co-design framework tailored for LLM inference on Hopper GPUs. By introducing FP8-aware kernel optimizations and RoPE-aware scaling strategies, \sysname substantially improves inference throughput and alleviates memory bandwidth bottlenecks. In future work, we aim to further enhance inference efficiency by exploring KV cache compression and paging techniques that leverage cache sparsity, yielding additional performance improvements for large-scale LLM development.


%% file: sections/07-appendix.tex
\newpage
\appendix


\section{Rationale for Choosing Scaling Factor $\sigma_{\text{FP8}}$}\label{ap:rationaleforscalingfactor}

In Section~\ref{subsubsec:int4fp8}, we explained how \sysname accelerates linear layer matrix multiplication using a mixed-precision quantization strategy. Selecting the FP8-precision scaling factor $\sigma_{\text{FP8}}$ for INT4 weight quantization is critical for maximizing the inference speed of quantized models. This choice optimally leverages the FP8 tensor cores available in the Hopper architecture, significantly enhancing throughput compared to state-of-the-art methods such as QServe~\citep{lin2024qserve}. Here, we provide a detailed comparison between \sysname and QServe in linear layer computation:

\begin{table}[htbp]
\centering
\caption{Linear layer matrix multiplication with mixed-quantization strategy}
\label{tab:rationale_sigma_fp8}
\resizebox{1.0\textwidth}{!}{%
\renewcommand{\arraystretch}{1.2}
\begin{tabular}{cccc|ccc}
\toprule
\textbf{} & \multicolumn{3}{c}{\textbf{Quantization Strategy}} & \multicolumn{3}{c}{\textbf{Implementation Details}} \\
\midrule
 & Weight & Scaling factor ($\sigma$) & Activation & Dequantization & GEMM + accum. & Re-quantization \\
\midrule
 \sysname & INT4 & \textbf{FP8} & FP8 & \makecell{\textbf{Lookup-table}} & FP8 tensor core & CUDA core \\
 \midrule
 QServe & INT4 & \makecell{BF16/INT8 \\ (two-level quant.)} & INT8 & \makecell{Weight unpacking + \\zero point substraction + \\ weight reordering}  & INT8 tensor core & CUDA core \\
\bottomrule
\end{tabular}%
}
\end{table}

As detailed in Table~\ref{tab:rationale_sigma_fp8}, the dequantization process in \sysname is much simpler than QServe. Specifically, \sysname utilizes an in-register lookup table for dequantization, eliminating the need for additional logical operations. In contrast, QServe requires multiple logical operations, including weight unpacking (unsigned INT4 to unsigned INT8 conversion), zero point subtraction (unsigned INT8 to signed INT8 conversion), and weight reordering to ensure register-level parallelism. This operational simplicity significantly contributes to the throughput improvements observed in feed-forward network layers, as shown in Figure~\ref{fig:int4-fp8-kernel}. The streamlined implementation and associated performance benefits underpin our choice of the $\sigma_{\text{FP8}}$ scaling factor.





\subsection{Detailed operational cost comparison}
We compare the quantization overhead of \sysname with QServe, a similar W4A8-based quantization method utilizing INT4 weights and INT8 activations. QServe initially performs per-channel quantization on the original BF16 weights $\mathbf{W_\text{BF16}}\in \mathbb{R}^{d_\text{out}\times d_\text{in}}$, with input dimension $d_\text{in}$ and output dimension $d_\text{out}$, converting them into INT8 weights $\mathbf{W_\text{INT8}}\in \mathbb{R}^{d_\text{out}\times d_\text{in}}$ using BF16 scale factors $\mathbf{s_\text{w}}\in \mathbb{R}^{d_\text{out}}$:
\begin{equation}
\mathbf{W_\text{BF16}}\simeq\mathbf{s_\text{w}}\mathbf{W_\text{INT8}}.
\end{equation}
It then applies group-wise quantization with group size $g$ to further convert INT8 weights into INT4 weights $\mathbf{W_\text{INT4}}\in \mathbb{R}^{d_\text{out}\times d_\text{in}}$, accompanied by corresponding scale factors and zero points $\mathbf{s_\text{INT8}},\mathbf{z_\text{INT4}} \in \mathbb{R}^{d_\text{out}\times d_\text{in}/g}$:
\begin{equation}
\mathbf{W_\text{INT8}}\simeq(\mathbf{W_\text{INT4}}-\mathbf{z_\text{INT4}})\cdot\mathbf{s_\text{INT8}}.
\end{equation}
Moreover, QServe employes per-token quantization for activations, transforming BF16 activations $\mathbf{x_\text{BF16}}\in \mathbb{R}^{b\times d_\text{in}}$, with batch size $b$, into INT8 activations $\mathbf{x_\text{INT8}}\in \mathbb{R}^{b\times d_\text{in}}$ using BF16 scale factor $\mathbf{s_x}\in \mathbb{R}^b$:
\begin{equation} \mathbf{x_\text{BF16}}\simeq\mathbf{s_\text{x}}\mathbf{x_\text{INT8}}.
\end{equation}
Consequently, QServe introduces significant overhead for dequantizing linear layer computations $\mathbf{x_\text{out}}=\mathbf{x_\text{in}}\mathbf{W}^T$, requiring $d_\text{out}\times d_\text{in}$ INT4 subtractions and INT8 multiplications for dequantizing weights into INT8, as well as $2\times b\times d_\text{out}$ BF16 multiplications for fused dequantization on the output activation.


In contrast, \sysname uses only a group-wise weight scale factor and per-token activation scale factor for quantization, leading to only $d_\text{out}\times d_\text{in}$ of FP8 register lookups, and $b\times d_\text{out}$ times of BF16 multiplications on the output activation. Therefore, compared to QServe, \sysname can reduce the computational overhead for dequantization operations in each linear layer to $(b+d_\text{in})\times d_\text{out}$.

\newpage
\section{Challenge: Underflow and overflow by $\sigma_{\text{FP8}}$}\label{ap:underflowandoverflowbyfp8}

\begin{figure}[h]
    \centering
    \includegraphics[width=0.6\linewidth]{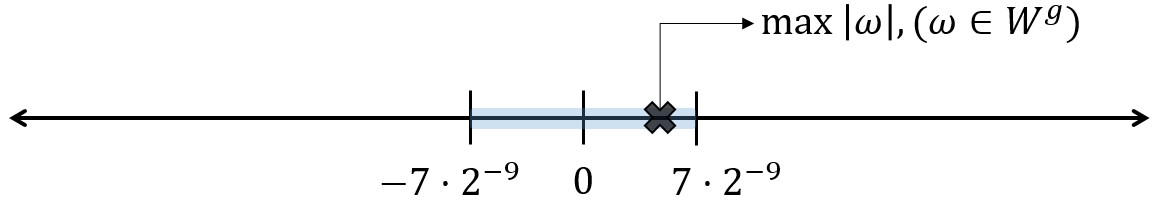}
    \caption{\small Example of underflow for a quantization group $W^g$ using $\sigma_{\text{FP8}}$.} 
    \label{fig:fp8_underflow}
\end{figure}

Although the FP8 scaling factor $\sigma_{\text{FP8}}$ substantially enhances throughput in quantized model serving, it can also lead to underflow and overflow issues due to FP8's limited dynamic range compared to BF16. Typically, underflow is more prevalent than overflow since the average magnitude of weight values is usually small and massive outliers rarely occur. Thus, we first address the underflow scenario. Figure~\ref{fig:fp8_underflow} provides an example of underflow in a quantization group $W^g \subset W$. 

\begin{lem}[\texttt{Underflow}]\label{lem:underflow}
    If $\max |\omega| < 7 \cdot 2^{-9}$, then all elements $\omega \in W^g$ in the quantization group are INT4-quantized to $0_{\text{INT4}}$ and subsquently FP8-dequantized to $0_{\text{FP8}}$ during the GEMM operation.
\end{lem}
\begin{proof}
    If $\max |\omega| < 7 \cdot 2^{-9},$ the scaling factor $\sigma_{\text{FP8}}$ is determined as follows:
    \begin{equation*}
        \sigma_{\text{FP8}} = \frac{\max|\omega|}{2^{(b-1)} -1} = \frac{\max|\omega|}{7}  < 2^{-9},
    \end{equation*}
    where $b$ represents the bit-width for INT4 quantization $(b = 4)$. Consequently, the scaling factor becomes zero:
    \begin{equation*}
        \sigma_{\text{FP8}} = 0,
    \end{equation*}
    as the smallest subnormal number representable by FP8 precision is $2^{-9}.$ Then, all elements within the quantization group are INT4-quantized to $0_{\text{INT4}}$ and subsequently FP8-dequantized in $0_{\text{FP8}}$ for the GEMM operation.
\end{proof}

\begin{figure}[h]
    \centering
    \includegraphics[width=0.6\linewidth]{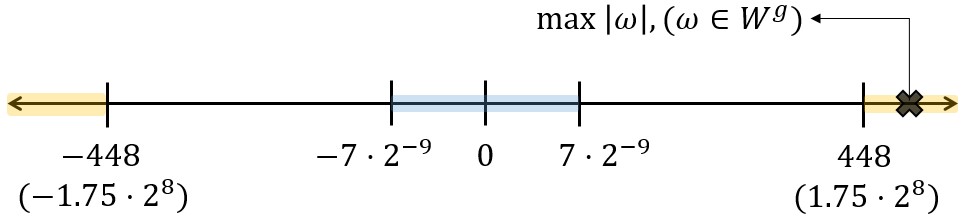}
    \caption{\small Example of overflow for a quantization group $W^g$ using $\sigma_{\text{FP8}}$.} 
    \label{fig:fp8_overflow}
\end{figure}

Furthermore, Figure~\ref{fig:fp8_overflow} illustrates a scenario in which overflow occurs. Although such overflow is rare, it can lead to accuracy degradation when it does occur. \sysname employs per-tensor scaling (PTS) primarily to prevent underflow; however, this approach may inadvertently cause overflow due to the upscaling of weight elements. Thus, identifying conditions under which overflow occurs is crucial. The following section explains how \sysname is designed to simultaneously mitigate underflow and prevent overflow.


\subsection{Discussion of $\sigma_{\text{FP8}}$ and $\sigma_{\text{FP16}}$}\label{ap:discussionBF16scalingfactor}

We previously discussed that while the FP8 scaling factor improves inference speed, it also triggers underflow and overflow issues that have a negative impact on accuracy. To comparatively illustrate the effects of FP8, we also included an evaluation using a BF16 scaling factor $\sigma_{\text{BF16}}$. As shown in Figure~\ref{fig:int4-fp8-kernel}, the feed-forward network layer using $\sigma_{\text{BF16}}$ exhibited a reduced speed, approximately 0.6x that of the $\sigma_{\text{FP8}}$ for Llama-3 8B model. However, Table~\ref{tab:zero-shot-accuracy} and~\ref{tab:ablation} show a slight increase in accuracy ($<0.4\%$) on both PPL and zero-shot accuracy benchmarks with $\sigma_{\text{BF16}}$. This result implies that adopting an FP8 scaling factor for INT4 quantization on the Hopper architecture achieves substantial speed improvements while incurring only minimal accuracy loss.

\newpage
\section{Mathematical Details}
\subsection{Discussion for Definition~\ref{def:per-tensor-scaling}: Per-Tensor Scaling (PTS)}\label{appedix:per-tensor-scaling}

We describe how per-tensor scaling, defined in Definition~\ref{def:per-tensor-scaling}, is formulated. The cumulative underflow score $\mathbb{S}(\mathbf{W})$ of weight matrix $\mathbf{W}$ is calculated by summing the distances of all elements $\omega \in \mathbf{W}$ that lie below the underflow threshold $7\cdot 2^{-9}$, as depicted in Figure~\ref{fig:fp8_underflow}. Elements above this threshold do not contribute to the score. Formally, the cumulative score $\mathbb{S}(\mathbf{W})$ is defined as:
\begin{equation*}
\mathbb{S}(\mathbf{W}) \triangleq \sum_{\omega \in \mathbf{W}} \max\left( 0,\; 7\cdot 2^{-9} - |\omega|\right).
\end{equation*}
A large score $\mathbb{S}(\mathbf{W})$ indicates many elements falling below the threshold. Hence, increasing the scaling factor $\delta$ reduces $\mathbb{S}(\mathbf{W})$. The first condition of PTS:
\begin{equation*}
\mathbb{S}(\mathbf{W}\cdot 2^n) = \mathbb{S}(\mathbf{W}\cdot 2^{n+i}),    
\end{equation*}
implies that \sysname chooses the minimum scaling factor $\delta=2^n,$ beyond which any larger scaling factor $\delta'= 2^{n+i} > \delta$ will not further reduce the cumulative underflow distance for any $i \in \mathbb{N}.$

The second condition of PTS is defined as follows:
\begin{equation*}
    7 \cdot 2^{5-n} \leq |\omega| < 7 \cdot 2^{6-n},
\end{equation*}
which indicates a risk of overflow if scaling goes beyond $\delta = 2^n$. As depicted in Figure~\ref{fig:fp8_overflow}, the overflow threshold is $1.75 \cdot 2^8$. Thus, the condition can be rewritten as: 
$$
1.75 \cdot 2^7\leq \delta \cdot |\omega| < 1.75 \cdot 2^8 \quad \text{or equivalently,} \quad 7 \cdot 2^{5-n} \leq |\omega| < 7\cdot 2^{6-n}.
$$
This means the sufficiently large scaling factor $2^n$ is selected such that no element $\omega \in \mathbf{W}$ exceeds the overflow threshold.


\subsection{Proof of Theorem~\ref{thm:rope_preserving_normalization}: RoPE-preserving Normalization (RPN) Bound}\label{ap:proofoftheorem}


\begin{proof}
    Since the shared scaling factor of channel $i$ and $j$ is defined by:
    \begin{equation*}
         s_j = s_i \triangleq \alpha \cdot \max_{n \in \{1, \dots, N\}} \left\| \left(k_i^n, k_j^n\right) \right\|_2,
    \end{equation*}
    for all $n \in \{1,\dots,N\},$ the following inequality holds:
    \begin{equation*}
        \frac{\left\| \left(k_i^n, k_j^n\right) \right\|_2}{\alpha \cdot   \max_{n \in \{1, \dots, N\}}\left\| \left(k_i^n, k_j^n\right) \right\|_2} \leq \frac{1}{\alpha}.
    \end{equation*}
\end{proof}

\newpage
\section{Performance of \sysname: Scaled-Dot Product Kernel in Attention Layer}\label{ap:scaled-dot-product}

\begin{figure}[h]
    \centering
    \includegraphics[width=0.7\linewidth]{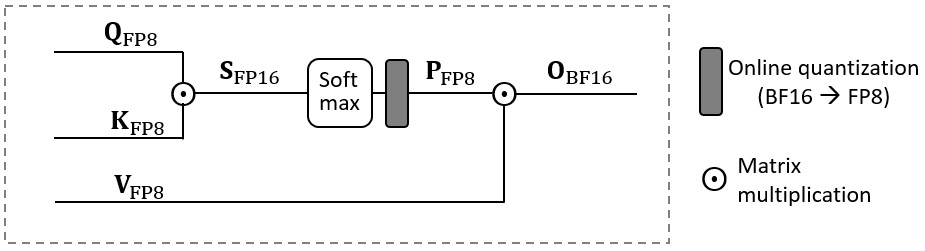}
    \caption{\small Scale-dot product kernel of \sysname.} 
    \label{fig:attention}
\end{figure}

Section~\ref{subsec:impact_of_prefill_phase_optimization} discussed the overall impact of prefill phase optimization on inference throughput. In this section, we specifically isolate and analyze the performance improvements by the scaled-dot product attention kernel, illustrated in Figure~\ref{fig:attention}. Note that the key and value matrices are quantized in FP8 (not in INT4) for the prefill phase. This is a core component of \sysname's three-stage pipelined optimization approach described in Algorithm~\ref{al:3-phase-fmha}.

\begin{figure}[h]
    \centering
    \includegraphics[width=0.8\linewidth]{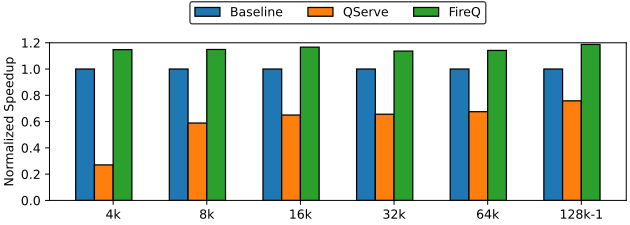}
    \caption{\small Comparison of scaled-dot product attention kernel performance between \sysname and QServe. The baseline is FlashAttention-2 BF16 kernel. We use the Llama-3.1 8B with varying input sequence length (x-axis).} 
    \label{fig:attention-performance}
\end{figure}

Performance evaluation results, depicted in Figure~\ref{fig:attention-performance}, demonstrate that \sysname significantly outperforms both the baseline FlashAttention-2 kernel (without quantization) and the comparative quantization framework, QServe. These results underscore the effectiveness of \sysname’s optimized three-stage pipeline in reducing latency and enhancing inference throughput across varying input sequence lengths for the Llama-3.1 (8B) model.


\newpage

\section{Ablation Study for Outlier Smoothing Strategies}\label{ap:ablation_for_smoothing_strategies}


\subsection{Impact of the Number of Outliers: RPN vs. CRS}

\begin{figure}[h]
\centering
    \subfloat[Perplexity without per-tensor scaling]{\includegraphics[width=0.4\linewidth]{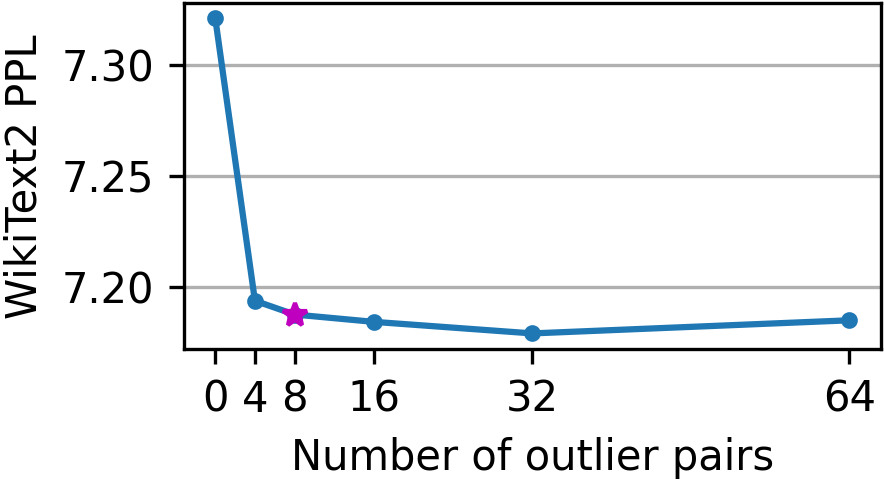}\label{fig:ppl_curve_wo_pts}}
    \hspace{0.5cm}
    \subfloat[Perplexity with per-tensor scaling]{\includegraphics[width=0.4\linewidth]{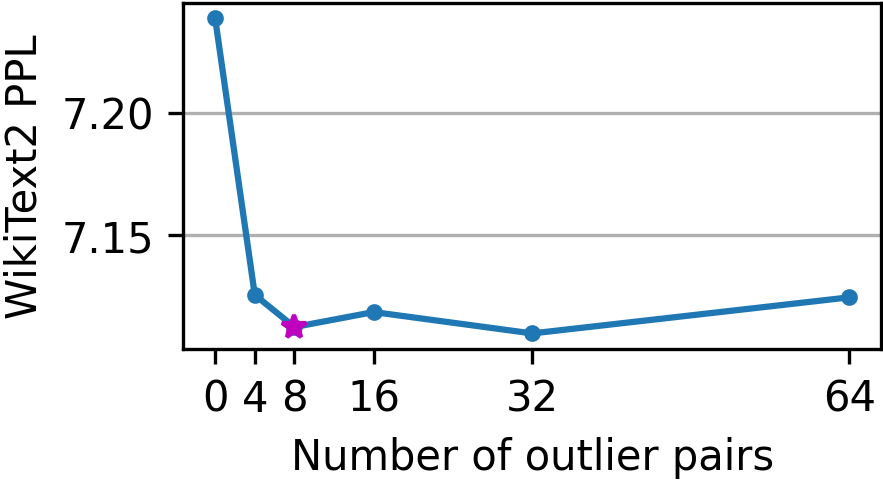}\label{fig:ppl_curve_w_pts}}
\caption{Perplexity on WikiText-2 for the Llama3.1 (8B) model. We select the shared scaling factor for RPN as $\alpha = 8$ and the scaling factor for CRS as $\beta = 8$. Per-tensor scaling is applied only in Figure~\ref{fig:ppl_curve_w_pts} (right).}
\label{fig:ppl_curve}
\end{figure}

As discussed in Section~\ref{subsubsec:rope_aware_quantization},
\sysname categorizes channels in the key matrix into two groups before applying RoPE-aware outlier smoothing: outlier channel pairs (handled by CRS) and normal channels (handled by RPN). Determining the optimal number of outlier channel pairs is crucial since it affects both model accuracy and inference speed, due to the differences in how CRS (computed online) and RPN (merged offline into weights) are implemented. Selecting too many outliers increases the computational runtime overhead from CRS, potentially reducing inference speed. Through calibration (see Figure~\ref{fig:ppl_curve}), we identified {\textbf{eight}} as the optimal number of outlier channel pairs. Beyond eight pairs, adding more outliers results in negligible accuracy gain. Thus, \sysname applies CRS to {\text{eight}} outlier channel pairs and RPN to all remaining channels.

\begin{figure}[h]
\centering
    \subfloat[Original]{\includegraphics[width=0.25\linewidth]{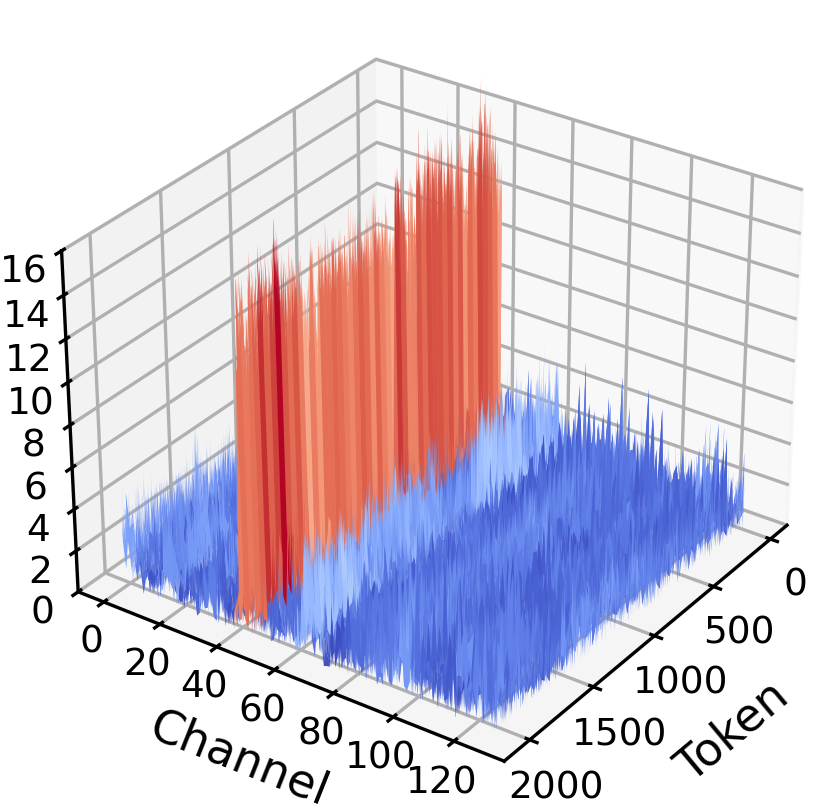}\label{fig:key_original}}
    \subfloat[64, RPN]{\includegraphics[width=0.25\linewidth]{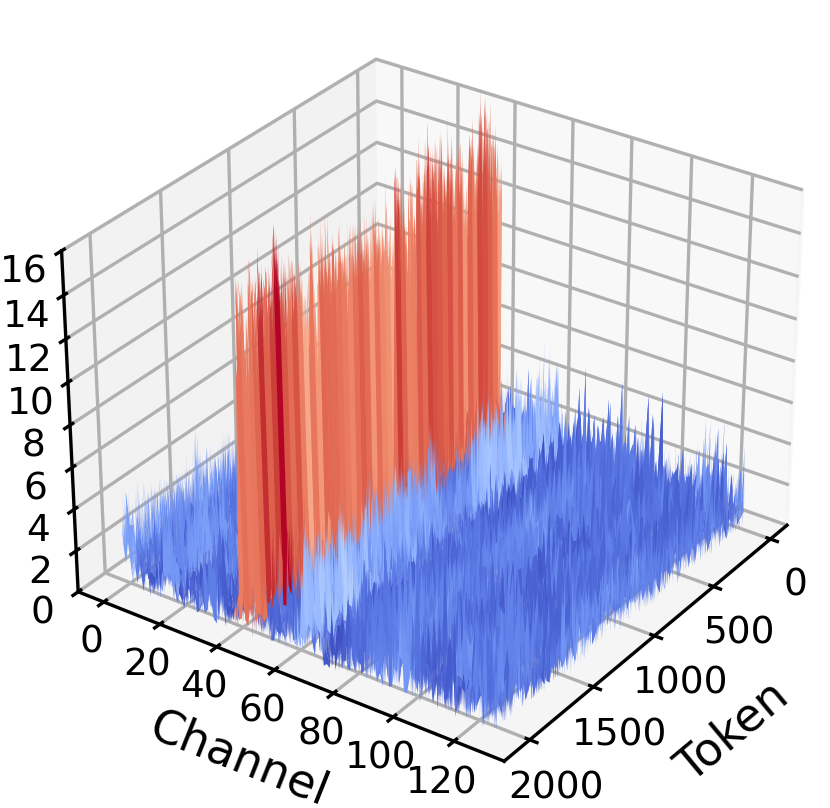}\label{fig:key_after_rpn_top64}}
    \subfloat[64, RPN+CRS]{\includegraphics[width=0.25\linewidth]{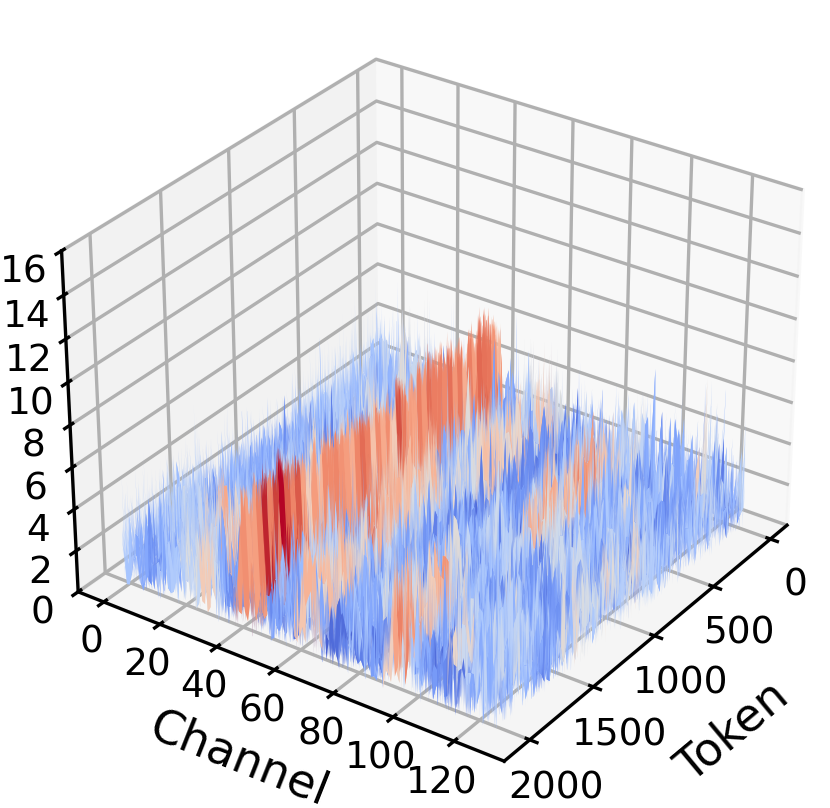}\label{fig:key_after_crs_top64}}
    \hfill
    \subfloat[0, RPN]{\includegraphics[width=0.25\linewidth]{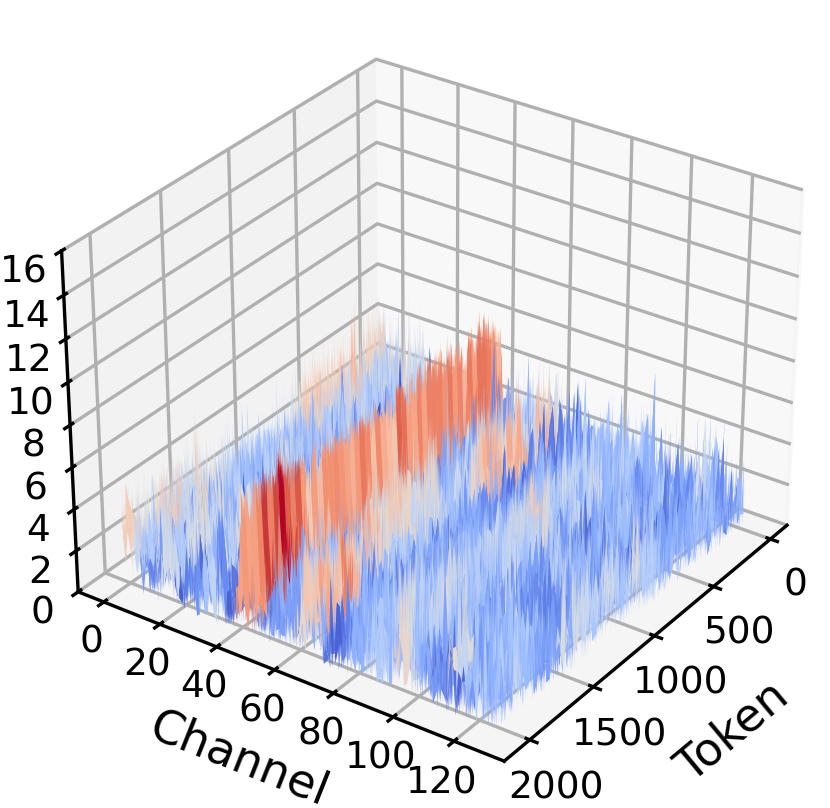}\label{fig:key_after_rpn_top0}}
    \subfloat[0, RPN+CRS]{\includegraphics[width=0.25\linewidth]{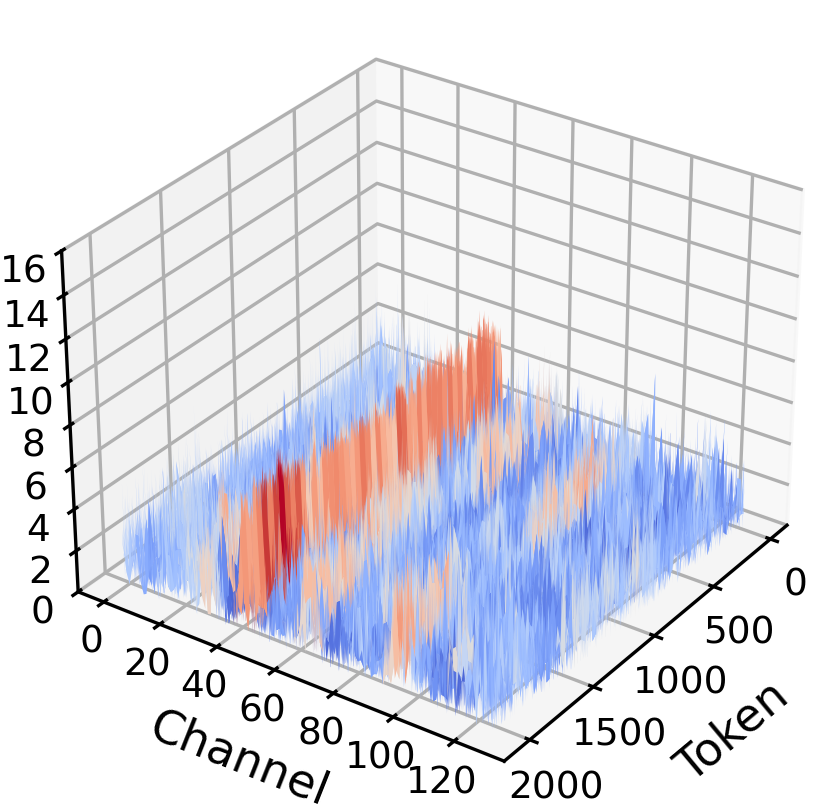}\label{fig:key_after_crs_top0}}
    \subfloat[8, RPN]{\includegraphics[width=0.25\linewidth]{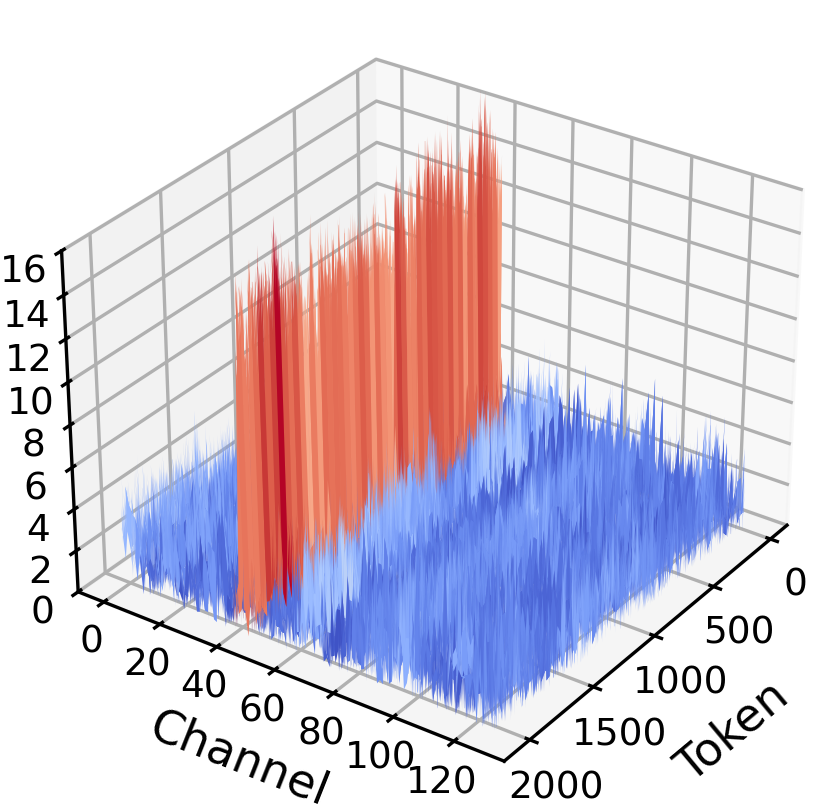}\label{fig:key_after_rpn_top8}}
    \subfloat[8, RPN+CRS]{\includegraphics[width=0.25\linewidth]{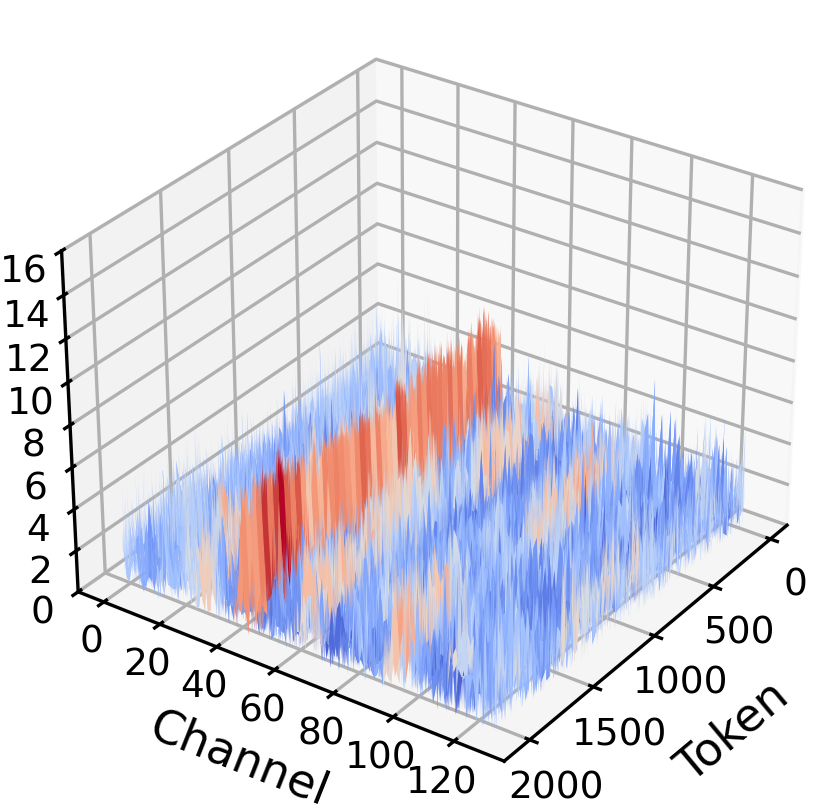}\label{fig:key_after_crs_top8}}

\caption{Number of outlier channel pairs, applied smoothing strategies (layer 25, head 6)}
\label{fig:ppl_curve}
\end{figure}

\subsection{Impact of RPN and CRS on Outlier Smoothing}

The key matrix typically contains outliers, as illustrated in Figure~\ref{fig:key_original}. When all channels (64 channels) are designated as outliers, the key matrix remains unchanged after RPN but is effectively smoothed after CRS, as depicted in Figure~\ref{fig:key_after_rpn_top64}. Conversely, Figure ~\ref{fig:key_after_rpn_top0} illustrates the scenario in which no outlier channels are selected, meaning all channels undergo RPN processing. Although the smoothing results from CRS-only (64 outlier channels, see Figure~\ref{fig:key_after_crs_top64}) and RPN-only (0 outlier channels, see Figure~\ref{fig:key_after_crs_top0}) cases appear visually similar, their performance differs; specifically, the PPL of the CRS-only case is better, as evident in the 64-channel pairs and 0-channel pairs in Figure~\ref{fig:ppl_curve}. Finally, when \sysname selects eight outlier channels, Figure~\ref{fig:key_after_rpn_top8} shows the key matrix after RPN, and Figure~\ref{fig:key_after_crs_top8} presents the results after applying both RPN and CRS. This approach effectively mitigates the impact of outliers.

\subsection{Impact of PTS and CAS on Underflow Mitigation}

\begin{figure}[h]
\centering
    \subfloat[Layer 0]{\includegraphics[width=0.42\linewidth]{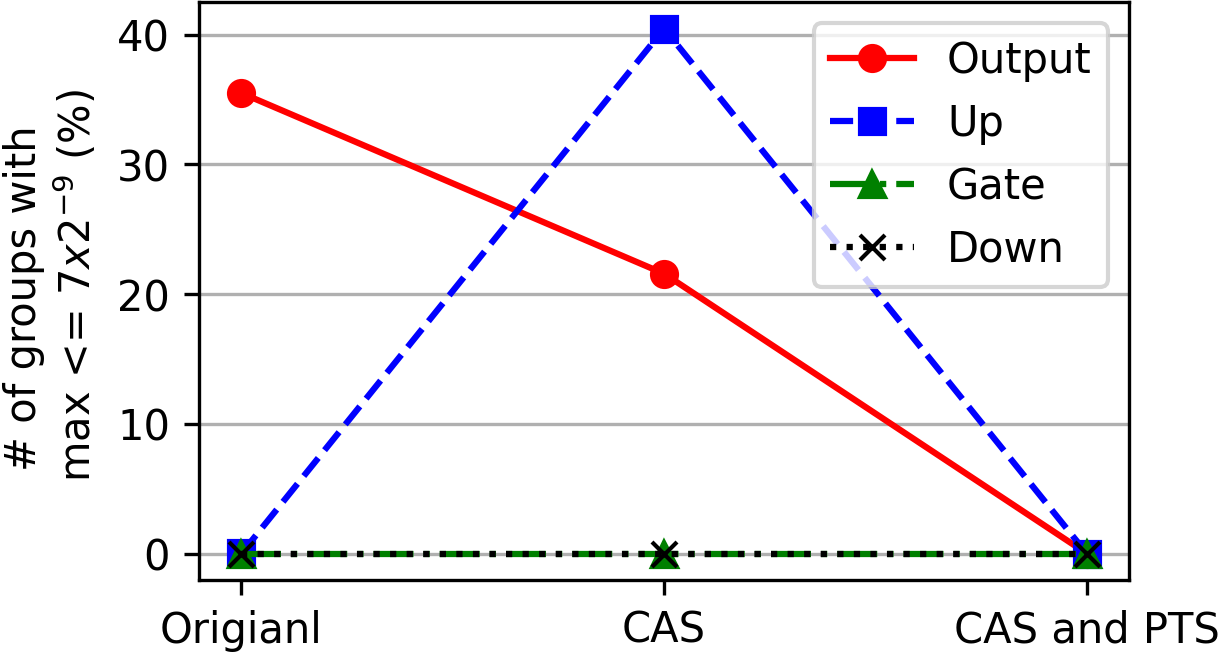}\label{fig:11}}
    \subfloat[Layer 10]{\includegraphics[width=0.42\linewidth]{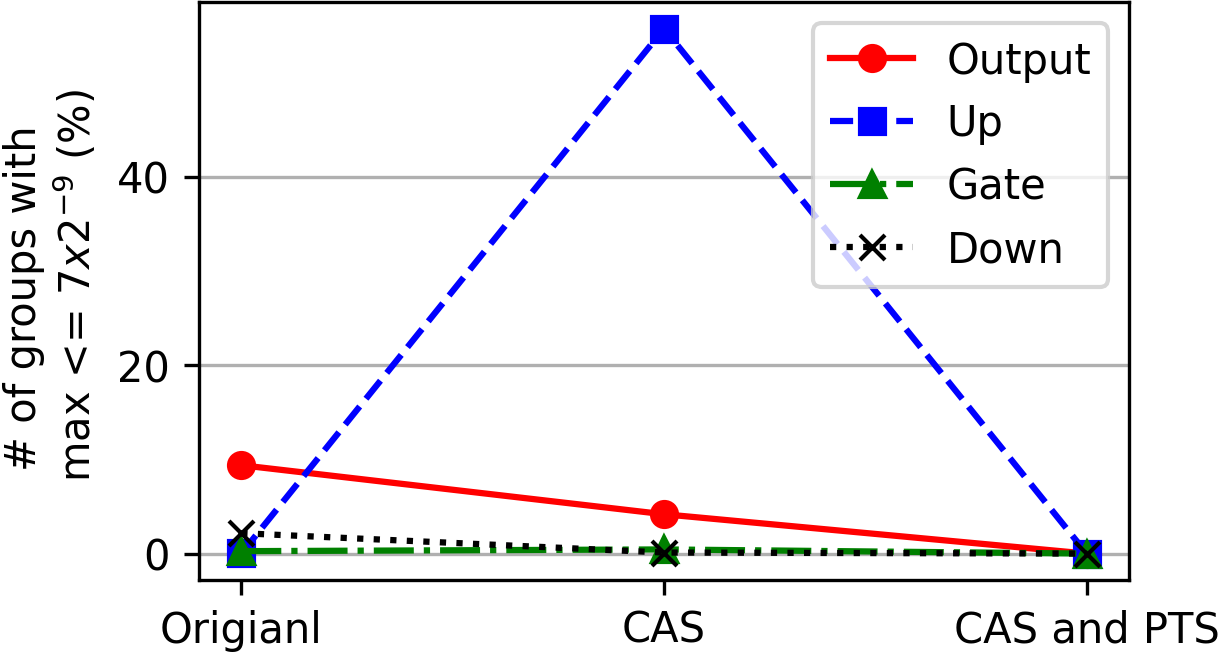}\label{fig:22}}
    \hfill
    \subfloat[Layer 20]{\includegraphics[width=0.42\linewidth]{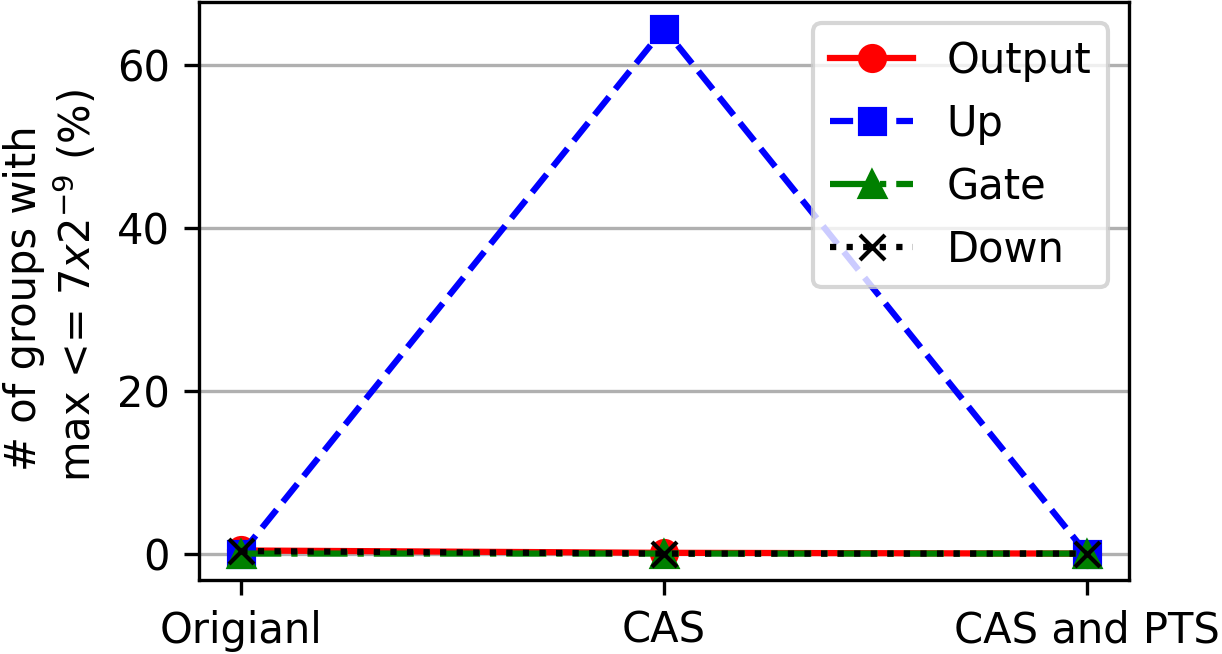}\label{fig:33}}
    \subfloat[Layer 30]{\includegraphics[width=0.42\linewidth]{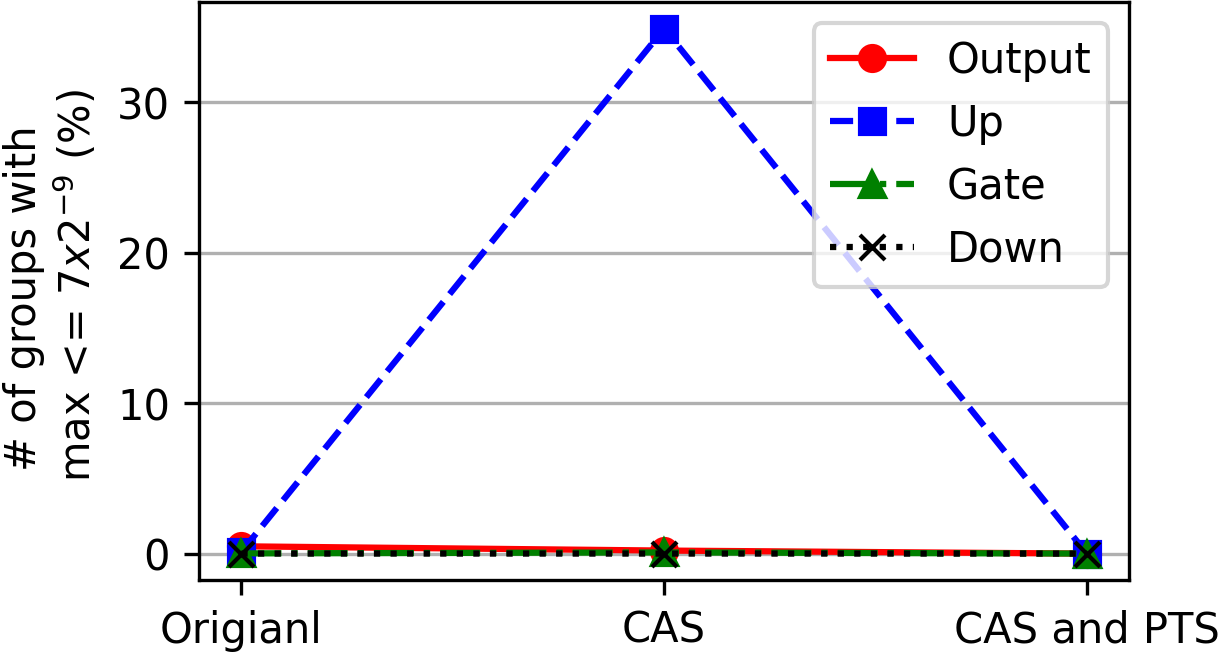}\label{fig:44}}
\caption{Impact of CAS and PTS on underflow mitigation for different layers.}
\label{fig:CASPTS}
\end{figure}

The per-tensor scaling (PTS) and channel-wise absmean scaling (CAS) strategies are designed to preserve accuracy during linear layer operations, as detailed in Section~\ref{subsubsec:linearlayerscaling}. CAS normalizes the distribution of each channel toward a shared target absolute mean, while PTS explicitly mitigates underflow by scaling up the weight matrix. Figure~\ref{fig:CASPTS} illustrates the effectiveness of CAS and PTS across various linear layers, including the up, gate, down, and out projections. The y-axis represents the percentage of quantization groups whose maximum elements fall below the FP8 underflow threshold of $7 \cdot 2^{-9}$, as derived in Appendix~\ref{ap:underflowandoverflowbyfp8}.


Figure~\ref{fig:CASPTS} further demonstrates the reduction of underflow-risk groups in gate, down, and out projections after applying CAS and PTS. Specifically, 
in the up projection, CAS initially normalizes the values close to the underflow threshold, after which PTS scales these elements upwards.
Together, these results confirm that the combination of CAS and PTS substantially mitigates underflow issues and effectively preserves accuracy.





\newpage
\section{Modified FlashAttention-3: prefill phase}\label{ap:full_prefill_phase}

Due to space constraints, we omitted the data preparation procedure for the \texttt{producer warpgroup} in Algorithm~\ref{al:3-phase-fmha}. Therefore, we provide the complete version of the attention later forward pass during the prefill phase here.

\begin{algorithm}[h]
\caption{Attention layer forward at prefill phase}
\label{al:3-phase-fmha-fullversion}
\small
\begin{spacing}{1.15}
\begin{algorithmic}[1]
\Require: Matrices $\mathbf{Q}_i\in \mathbb{R}^{B_r \times d}, \mathbf{K}, \mathbf{V} \in \mathbb{R}^{N \times d}$ in \texttt{GMEM}, query block size $B_r$, key and value block size $B_c$ with $T_c=\lceil N/B_c \rceil$
\State Initialize pipeline objects to manage barrier synchronization with a 2-stage \texttt{SMEM} buffer
\State Fetch the query index $i$ from global memory with atomic decrement
\If{\texttt{Producer warpgroup}}
    \State Deallocate the predetermined number of registers
    \State TMA load $\mathbf{Q}_{i0}$, $\mathbf{K}_0$, $\mathbf{Q}_{i1}$, $\mathbf{K}_1$ from \texttt{GMEM} to \texttt{SMEM} in order
    \For{$2\leq j < T_c$}
        \State TMA load $\mathbf{V}_{j-2}$ from \texttt{GMEM} to \texttt{SMEM} and transpose to $\mathbf{V}_{j-2}^\text{T}$
        \State TMA load $\textbf{K}_j$ from \texttt{GMEM} to \texttt{SMEM}
    \EndFor
    \State TMA load $\mathbf{V}_{T_c-2}$ from \texttt{GMEM} to \texttt{SMEM} and transpose to $\mathbf{V}_{T_c-2}^\text{T}$
    \State TMA load $\mathbf{V}_{T_c-1}$ from \texttt{GMEM} to \texttt{SMEM} and transpose to $\mathbf{V}_{T_c-1}^\text{T}$
\ElsIf{\texttt{Consumer warpgroup}}
    \State Allocate the predetermined number of registers
    \State Initialize $\mathbf{O}_{ik}=(0)\in \mathbb{R}^{B_r\times d}$ for $k \in \{0,1\}$, 
    ~$s_i,l_i,m_i=(1),(0),(-\infty)\in\mathbb{R}^{B_r}$
    \State Commit $\textbf{S}_{i0}=\texttt{wgmma}\_\texttt{ss}(\mathbf{Q}_{ik}, \mathbf{K}_0)$ and \texttt{wait} $\textbf{S}_{i0}$
    \State Commit $\textbf{S}_{i1}=\texttt{wgmma}\_\texttt{ss}(\mathbf{Q}_{ik}, \mathbf{K}_1)$ and do not \texttt{wait} $\textbf{S}_{i1}$
    \State Compute $\textbf{S}_{i0}=\text{mask}(\tau \textbf{S}_{i0})$, $m_i=\max\left(\text{rowmax}(S_{i0})\right), ~\textbf{P}_{i0}=\exp(\textbf{S}_{i0}-m_i),~l_i=\text{rowsum}(\textbf{P}_{i0})$
        \State \texttt{Wait} $\textbf{S}_{i1}$
    \For{$2\leq j < T_c$}
        \State Commit $\mathbf{O}_{ik}=\mathbf{O}_{ik}+\texttt{wgmma}\_\text{rs}\left(\textbf{P}_{i(j-2)}, \mathbf{V}_{j-2}^\text{T}\right)$ and do not \texttt{wait} $\mathbf{O}_{ik}$
        \State Compute $\textbf{S}_{i(j-1)}=\text{mask}(\tau \textbf{S}_{i(j-1)})$
        \State $m_{\text{old}}=m_i$ and compute $m_i=\max\left(m^{\text{old}}_i, \text{rowmax}(\textbf{S}_{i(j-1)})\right),~s_i=\exp(m^{\text{old}}_i-m_i), ~l_i=s_i\odot l_i$
        \State Commit $\textbf{S}_{ij}=\texttt{wgmma}\_\texttt{ss}(\textbf{Q}_{ik}, \mathbf{K}_j)$ and do not \texttt{wait} $\textbf{S}_{ij}$
        \State Compute $\textbf{P}_{i(j-1)}=\exp(\textbf{S}_{i(j-1)}-m_i),~l_i=l_i+\text{rowsum}(\textbf{P}_{i(j-1)})$
        \State \texttt{Wait} $\mathbf{O}_{ik}$ and compute $\mathbf{O}_{ik}=s_i\odot \mathbf{O}_{ik}$
        \State \texttt{Wait} $\textbf{S}_{ij}$
    \EndFor
    \State Commit $\mathbf{O}_{ik}=\mathbf{O}_{ik}+\texttt{wgmma}\_\texttt{rs}\left(\textbf{P}_{i(T_c-2)}, \mathbf{V}_{T_c-2}^\text{T}\right)$ and do not \texttt{wait} $\mathbf{O}_{ik}$
    \State $m^{\text{old}}_i=m_i$ and compute $m_i=\max\left(m^{\text{old}}_i, \text{rowmax}(\textbf{S}_{i(T_c-1)})\right),~s_i=\exp(m^{\text{old}}_i-m_i), ~l_i=s_i\odot l_i$
    \State Compute $\textbf{P}_{i(j-1)}=\exp(\textbf{S}_{i(T_c-1)}-m_i),~l_i=l_i+\text{rowsum}(\textbf{P}_{i(T_c-1)})$
    \State \texttt{Wait} $\mathbf{O}_{ik}$ and compute $\mathbf{O}_{ik}=\mathbf{s}_i\odot \mathbf{O}_{ik}$
    \State Commit $\mathbf{O}_{ik}=\mathbf{O}_{ik}+\texttt{wgmma}\_\texttt{rs}\left(\mathbf{P}_{i(T_c-1)}, \mathbf{V}_{T_c-1}^\text{T}\right)$ and \texttt{wait} $\mathbf{O}_{ik}$
    \State Compute $\mathbf{O}_{ik}=1/l_i \odot \mathbf{O}_{ik}$
    \State TMA store $\mathbf{O}_{ik}$ from \texttt{SMEM} to \texttt{GMEM}
\EndIf
\end{algorithmic}
\end{spacing}
\end{algorithm}

%% file: main.bbl

\begin{thebibliography}{23}


\ifx \showCODEN    \undefined \def \showCODEN     #1{\unskip}     \fi
\ifx \showDOI      \undefined \def \showDOI       #1{#1}\fi
\ifx \showISBNx    \undefined \def \showISBNx     #1{\unskip}     \fi
\ifx \showISBNxiii \undefined \def \showISBNxiii  #1{\unskip}     \fi
\ifx \showISSN     \undefined \def \showISSN      #1{\unskip}     \fi
\ifx \showLCCN     \undefined \def \showLCCN      #1{\unskip}     \fi
\ifx \shownote     \undefined \def \shownote      #1{#1}          \fi
\ifx \showarticletitle \undefined \def \showarticletitle #1{#1}   \fi
\ifx \showURL      \undefined \def \showURL       {\relax}        \fi
\providecommand\bibfield[2]{#2}
\providecommand\bibinfo[2]{#2}
\providecommand\natexlab[1]{#1}
\providecommand\showeprint[2][]{arXiv:#2}

\bibitem[AI@Meta(2024)]%
        {llama3}
\bibfield{author}{\bibinfo{person}{AI@Meta}.} \bibinfo{year}{2024}\natexlab{}.
\newblock \bibinfo{title}{Llama 3 model card}.
\newblock
\newblock
\urldef\tempurl%
\url{https://github.com/meta-llama/llama3/blob/main/MODEL_CARD.md}
\showURL{%
\tempurl}


\bibitem[Ashkboos et~al\mbox{.}(2024)]%
        {ashkboos2024quarot}
\bibfield{author}{\bibinfo{person}{Saleh Ashkboos}, \bibinfo{person}{Amirkeivan Mohtashami}, \bibinfo{person}{Maximilian Croci}, \bibinfo{person}{Bo Li}, \bibinfo{person}{Pashmina Cameron}, \bibinfo{person}{Martin Jaggi}, \bibinfo{person}{Dan Alistarh}, \bibinfo{person}{Torsten Hoefler}, {and} \bibinfo{person}{James Hensman}.} \bibinfo{year}{2024}\natexlab{}.
\newblock \showarticletitle{Quarot: Outlier-free 4-bit inference in rotated llms}.
\newblock \bibinfo{journal}{\emph{Advances in Neural Information Processing Systems}}  \bibinfo{volume}{37} (\bibinfo{year}{2024}), \bibinfo{pages}{100213--100240}.
\newblock


\bibitem[Bisk et~al\mbox{.}(2020)]%
        {bisk2020piqa}
\bibfield{author}{\bibinfo{person}{Yonatan Bisk}, \bibinfo{person}{Rowan Zellers}, \bibinfo{person}{Jianfeng Gao}, \bibinfo{person}{Yejin Choi}, {et~al\mbox{.}}} \bibinfo{year}{2020}\natexlab{}.
\newblock \showarticletitle{Piqa: Reasoning about physical commonsense in natural language}. In \bibinfo{booktitle}{\emph{Proceedings of the AAAI conference on artificial intelligence}}, Vol.~\bibinfo{volume}{34}. \bibinfo{pages}{7432--7439}.
\newblock


\bibitem[Clark et~al\mbox{.}(2019)]%
        {clark2019boolq}
\bibfield{author}{\bibinfo{person}{Christopher Clark}, \bibinfo{person}{Kenton Lee}, \bibinfo{person}{Ming-Wei Chang}, \bibinfo{person}{Tom Kwiatkowski}, \bibinfo{person}{Michael Collins}, {and} \bibinfo{person}{Kristina Toutanova}.} \bibinfo{year}{2019}\natexlab{}.
\newblock \showarticletitle{Boolq: Exploring the surprising difficulty of natural yes/no questions}.
\newblock \bibinfo{journal}{\emph{arXiv preprint arXiv:1905.10044}} (\bibinfo{year}{2019}).
\newblock


\bibitem[Clark et~al\mbox{.}(2018)]%
        {clark2018thinkarc}
\bibfield{author}{\bibinfo{person}{Peter Clark}, \bibinfo{person}{Isaac Cowhey}, \bibinfo{person}{Oren Etzioni}, \bibinfo{person}{Tushar Khot}, \bibinfo{person}{Ashish Sabharwal}, \bibinfo{person}{Carissa Schoenick}, {and} \bibinfo{person}{Oyvind Tafjord}.} \bibinfo{year}{2018}\natexlab{}.
\newblock \showarticletitle{Think you have solved question answering? try arc, the ai2 reasoning challenge}.
\newblock \bibinfo{journal}{\emph{arXiv preprint arXiv:1803.05457}} (\bibinfo{year}{2018}).
\newblock


\bibitem[Gholami et~al\mbox{.}(2024)]%
        {memory-wall}
\bibfield{author}{\bibinfo{person}{Amir Gholami}, \bibinfo{person}{Zhewei Yao}, \bibinfo{person}{Sehoon Kim}, \bibinfo{person}{Coleman Hooper}, \bibinfo{person}{Michael~W. Mahoney}, {and} \bibinfo{person}{Kurt Keutzer}.} \bibinfo{year}{2024}\natexlab{}.
\newblock \showarticletitle{AI and Memory Wall}.
\newblock \bibinfo{journal}{\emph{IEEE Micro}} \bibinfo{volume}{44}, \bibinfo{number}{3} (\bibinfo{year}{2024}), \bibinfo{pages}{33--39}.
\newblock
\urldef\tempurl%
\url{https://doi.org/10.1109/MM.2024.3373763}
\showDOI{\tempurl}


\bibitem[Hooper et~al\mbox{.}(2024)]%
        {hooper2024kvquant}
\bibfield{author}{\bibinfo{person}{Coleman Hooper}, \bibinfo{person}{Sehoon Kim}, \bibinfo{person}{Hiva Mohammadzadeh}, \bibinfo{person}{Michael~W Mahoney}, \bibinfo{person}{Sophia Shao}, \bibinfo{person}{Kurt Keutzer}, {and} \bibinfo{person}{Amir Gholami}.} \bibinfo{year}{2024}\natexlab{}.
\newblock \showarticletitle{Kvquant: Towards 10 million context length llm inference with kv cache quantization}.
\newblock \bibinfo{journal}{\emph{Advances in Neural Information Processing Systems}}  \bibinfo{volume}{37} (\bibinfo{year}{2024}), \bibinfo{pages}{1270--1303}.
\newblock


\bibitem[Kwon et~al\mbox{.}(2023)]%
        {vllm}
\bibfield{author}{\bibinfo{person}{Woosuk Kwon}, \bibinfo{person}{Zhuohan Li}, \bibinfo{person}{Siyuan Zhuang}, \bibinfo{person}{Ying Sheng}, \bibinfo{person}{Lianmin Zheng}, \bibinfo{person}{Cody~Hao Yu}, \bibinfo{person}{Joseph~E. Gonzalez}, \bibinfo{person}{Hao Zhang}, {and} \bibinfo{person}{Ion Stoica}.} \bibinfo{year}{2023}\natexlab{}.
\newblock \showarticletitle{Efficient Memory Management for Large Language Model Serving with PagedAttention}. In \bibinfo{booktitle}{\emph{Proceedings of the ACM SIGOPS 29th Symposium on Operating Systems Principles}}.
\newblock


\bibitem[Lee et~al\mbox{.}(2024)]%
        {skhynix-aim}
\bibfield{author}{\bibinfo{person}{Hyungdeok Lee}, \bibinfo{person}{Guhyun Kim}, \bibinfo{person}{Dayeon Yun}, \bibinfo{person}{Ilkon Kim}, \bibinfo{person}{Yongkee Kwon}, {and} \bibinfo{person}{Euicheol Lim}.} \bibinfo{year}{2024}\natexlab{}.
\newblock \showarticletitle{Cost-Effective LLM Accelerator Using Processing in Memory Technology}. In \bibinfo{booktitle}{\emph{2024 IEEE Symposium on VLSI Technology and Circuits (VLSI Technology and Circuits)}}. \bibinfo{pages}{1--2}.
\newblock
\urldef\tempurl%
\url{https://doi.org/10.1109/VLSITechnologyandCir46783.2024.10631397}
\showDOI{\tempurl}


\bibitem[Lin et~al\mbox{.}(2024a)]%
        {lin2024awq}
\bibfield{author}{\bibinfo{person}{Ji Lin}, \bibinfo{person}{Jiaming Tang}, \bibinfo{person}{Haotian Tang}, \bibinfo{person}{Shang Yang}, \bibinfo{person}{Wei-Ming Chen}, \bibinfo{person}{Wei-Chen Wang}, \bibinfo{person}{Guangxuan Xiao}, \bibinfo{person}{Xingyu Dang}, \bibinfo{person}{Chuang Gan}, {and} \bibinfo{person}{Song Han}.} \bibinfo{year}{2024}\natexlab{a}.
\newblock \showarticletitle{Awq: Activation-aware weight quantization for on-device llm compression and acceleration}.
\newblock \bibinfo{journal}{\emph{Proceedings of Machine Learning and Systems}}  \bibinfo{volume}{6} (\bibinfo{year}{2024}), \bibinfo{pages}{87--100}.
\newblock


\bibitem[Lin et~al\mbox{.}(2024b)]%
        {lin2024qserve}
\bibfield{author}{\bibinfo{person}{Yujun Lin}, \bibinfo{person}{Haotian Tang}, \bibinfo{person}{Shang Yang}, \bibinfo{person}{Zhekai Zhang}, \bibinfo{person}{Guangxuan Xiao}, \bibinfo{person}{Chuang Gan}, {and} \bibinfo{person}{Song Han}.} \bibinfo{year}{2024}\natexlab{b}.
\newblock \showarticletitle{Qserve: W4a8kv4 quantization and system co-design for efficient llm serving}.
\newblock \bibinfo{journal}{\emph{arXiv preprint arXiv:2405.04532}} (\bibinfo{year}{2024}).
\newblock


\bibitem[Liu et~al\mbox{.}(2024)]%
        {liu2024spinquant}
\bibfield{author}{\bibinfo{person}{Zechun Liu}, \bibinfo{person}{Changsheng Zhao}, \bibinfo{person}{Igor Fedorov}, \bibinfo{person}{Bilge Soran}, \bibinfo{person}{Dhruv Choudhary}, \bibinfo{person}{Raghuraman Krishnamoorthi}, \bibinfo{person}{Vikas Chandra}, \bibinfo{person}{Yuandong Tian}, {and} \bibinfo{person}{Tijmen Blankevoort}.} \bibinfo{year}{2024}\natexlab{}.
\newblock \showarticletitle{Spinquant: Llm quantization with learned rotations}.
\newblock \bibinfo{journal}{\emph{arXiv preprint arXiv:2405.16406}} (\bibinfo{year}{2024}).
\newblock


\bibitem[Merity et~al\mbox{.}(2016)]%
        {merity2016pointerwikitext}
\bibfield{author}{\bibinfo{person}{Stephen Merity}, \bibinfo{person}{Caiming Xiong}, \bibinfo{person}{James Bradbury}, {and} \bibinfo{person}{Richard Socher}.} \bibinfo{year}{2016}\natexlab{}.
\newblock \showarticletitle{Pointer sentinel mixture models}.
\newblock \bibinfo{journal}{\emph{arXiv preprint arXiv:1609.07843}} (\bibinfo{year}{2016}).
\newblock


\bibitem[NVIDIA(2017)]%
        {cutlass}
\bibfield{author}{\bibinfo{person}{NVIDIA}.} \bibinfo{year}{2017}\natexlab{}.
\newblock \bibinfo{title}{CUDA Templates for Linear Algebra Subroutines}.
\newblock
\newblock
\urldef\tempurl%
\url{https://github.com/NVIDIA/cutlass}
\showURL{%
\tempurl}


\bibitem[NVIDIA(2023)]%
        {tensorrt-llm}
\bibfield{author}{\bibinfo{person}{NVIDIA}.} \bibinfo{year}{2023}\natexlab{}.
\newblock \bibinfo{title}{TensorRT-LLM: A TensorRT Toolbox for Optimized Large Language Model Inference}.
\newblock
\newblock
\urldef\tempurl%
\url{https://github.com/NVIDIA/TensorRT-LLM}
\showURL{%
\tempurl}


\bibitem[Sakaguchi et~al\mbox{.}(2021)]%
        {sakaguchi2021winogrande}
\bibfield{author}{\bibinfo{person}{Keisuke Sakaguchi}, \bibinfo{person}{Ronan~Le Bras}, \bibinfo{person}{Chandra Bhagavatula}, {and} \bibinfo{person}{Yejin Choi}.} \bibinfo{year}{2021}\natexlab{}.
\newblock \showarticletitle{Winogrande: An adversarial winograd schema challenge at scale}.
\newblock \bibinfo{journal}{\emph{Commun. ACM}} \bibinfo{volume}{64}, \bibinfo{number}{9} (\bibinfo{year}{2021}), \bibinfo{pages}{99--106}.
\newblock


\bibitem[Shah et~al\mbox{.}(2024)]%
        {shah2024flashattention}
\bibfield{author}{\bibinfo{person}{Jay Shah}, \bibinfo{person}{Ganesh Bikshandi}, \bibinfo{person}{Ying Zhang}, \bibinfo{person}{Vijay Thakkar}, \bibinfo{person}{Pradeep Ramani}, {and} \bibinfo{person}{Tri Dao}.} \bibinfo{year}{2024}\natexlab{}.
\newblock \showarticletitle{Flashattention-3: Fast and accurate attention with asynchrony and low-precision}.
\newblock \bibinfo{journal}{\emph{Advances in Neural Information Processing Systems}}  \bibinfo{volume}{37} (\bibinfo{year}{2024}), \bibinfo{pages}{68658--68685}.
\newblock


\bibitem[Su et~al\mbox{.}(2021)]%
        {su2021roformer}
\bibfield{author}{\bibinfo{person}{Jianlin Su}, \bibinfo{person}{Yu Lu}, \bibinfo{person}{Shengfeng Pan}, \bibinfo{person}{Bo Wen}, {and} \bibinfo{person}{Yunfeng Liu}.} \bibinfo{year}{2021}\natexlab{}.
\newblock \bibinfo{title}{RoFormer: Enhanced Transformer with Rotary Position Embedding}.
\newblock
\newblock
\showeprint[arxiv]{2104.09864}~[cs.CL]


\bibitem[Su et~al\mbox{.}(2025)]%
        {su2025rotatekv}
\bibfield{author}{\bibinfo{person}{Zunhai Su}, \bibinfo{person}{Zhe Chen}, \bibinfo{person}{Wang Shen}, \bibinfo{person}{Hanyu Wei}, \bibinfo{person}{Linge Li}, \bibinfo{person}{Huangqi Yu}, {and} \bibinfo{person}{Kehong Yuan}.} \bibinfo{year}{2025}\natexlab{}.
\newblock \showarticletitle{RotateKV: Accurate and Robust 2-Bit KV Cache Quantization for LLMs via Outlier-Aware Adaptive Rotations}.
\newblock \bibinfo{journal}{\emph{arXiv preprint arXiv:2501.16383}} (\bibinfo{year}{2025}).
\newblock


\bibitem[Touvron et~al\mbox{.}(2023)]%
        {touvron2023llama2}
\bibfield{author}{\bibinfo{person}{Hugo Touvron}, \bibinfo{person}{Louis Martin}, \bibinfo{person}{Kevin Stone}, \bibinfo{person}{Peter Albert}, \bibinfo{person}{Amjad Almahairi}, \bibinfo{person}{Yasmine Babaei}, \bibinfo{person}{Nikolay Bashlykov}, \bibinfo{person}{Soumya Batra}, \bibinfo{person}{Prajjwal Bhargava}, \bibinfo{person}{Shruti Bhosale}, {et~al\mbox{.}}} \bibinfo{year}{2023}\natexlab{}.
\newblock \showarticletitle{Llama 2: Open foundation and fine-tuned chat models}.
\newblock \bibinfo{journal}{\emph{arXiv preprint arXiv:2307.09288}} (\bibinfo{year}{2023}).
\newblock


\bibitem[Xiao et~al\mbox{.}(2023)]%
        {xiao2023smoothquant}
\bibfield{author}{\bibinfo{person}{Guangxuan Xiao}, \bibinfo{person}{Ji Lin}, \bibinfo{person}{Mickael Seznec}, \bibinfo{person}{Hao Wu}, \bibinfo{person}{Julien Demouth}, {and} \bibinfo{person}{Song Han}.} \bibinfo{year}{2023}\natexlab{}.
\newblock \showarticletitle{Smoothquant: Accurate and efficient post-training quantization for large language models}. In \bibinfo{booktitle}{\emph{International Conference on Machine Learning}}. PMLR, \bibinfo{pages}{38087--38099}.
\newblock


\bibitem[Zellers et~al\mbox{.}(2019)]%
        {zellers2019hellaswag}
\bibfield{author}{\bibinfo{person}{Rowan Zellers}, \bibinfo{person}{Ari Holtzman}, \bibinfo{person}{Yonatan Bisk}, \bibinfo{person}{Ali Farhadi}, {and} \bibinfo{person}{Yejin Choi}.} \bibinfo{year}{2019}\natexlab{}.
\newblock \showarticletitle{Hellaswag: Can a machine really finish your sentence?}
\newblock \bibinfo{journal}{\emph{arXiv preprint arXiv:1905.07830}} (\bibinfo{year}{2019}).
\newblock


\bibitem[Zhao et~al\mbox{.}(2024)]%
        {zhao2024atom}
\bibfield{author}{\bibinfo{person}{Yilong Zhao}, \bibinfo{person}{Chien-Yu Lin}, \bibinfo{person}{Kan Zhu}, \bibinfo{person}{Zihao Ye}, \bibinfo{person}{Lequn Chen}, \bibinfo{person}{Size Zheng}, \bibinfo{person}{Luis Ceze}, \bibinfo{person}{Arvind Krishnamurthy}, \bibinfo{person}{Tianqi Chen}, {and} \bibinfo{person}{Baris Kasikci}.} \bibinfo{year}{2024}\natexlab{}.
\newblock \showarticletitle{Atom: Low-bit quantization for efficient and accurate llm serving}.
\newblock \bibinfo{journal}{\emph{Proceedings of Machine Learning and Systems}}  \bibinfo{volume}{6} (\bibinfo{year}{2024}), \bibinfo{pages}{196--209}.
\newblock


\end{thebibliography}
